\documentclass[]{article}
\usepackage{jmlr2e}

\usepackage{times}

\usepackage{graphicx} 

\usepackage{bm}
\usepackage{amsmath}

\usepackage{algorithm}
\usepackage{algorithmic}

\date{}

\newcommand{\lp}{\left(}
\newcommand{\rp}{\right)}

\newcommand{\eps}{\epsilon}

\newcommand{\xs}{x}
\newcommand{\x}{{\bf x}}
\newcommand{\y}{{\bf y}}

\newcommand{\xps}{\x^{\prime(s)}}

\newcommand{\thetav}{{\bm \theta}}
\newcommand{\thetapv}{{\bm \theta^{\prime}}}
\newcommand{\thetavp}{\thetapv}

\newcommand{\eye}{{\bm I}}

\newcommand{\simsim}{\overset{\simulator}{\sim}}

\newcommand{\thetastar}{\thetav^{\star}}

\newcommand{\epsvec}{ {\bm \eps}}

\DeclareMathOperator\simulator{sim}

\newcommand{\gradv}{ {\bm g }}
\newcommand{\gradvest}{ \hat{\gradv}}

\newcommand{\rhov}{ {\bm \rho}}
\newcommand{\zerov}{ {\bm 0}}
\newcommand{\omegav}{ {\bm \omega}}

\newcommand{\varv}{ {\bm V}}
\newcommand{\Bv}{ {\bm B}}
\newcommand{\Cv}{ {\bm C}}
\newcommand{\yi}{\y_i}
\newcommand{\Deltav}{{\bm \Delta}}
\newcommand{\Deltavd}{\Deltav_d}
\newcommand{\Deltavr}{\Deltav_r}

\newcommand{\lleps}{\mathcal{L}_{\epsvec}}
\newcommand{\dtheta}{d_{\thetav}}

\title{Hamiltonian ABC}

\ShortHeadings{Hamiltonian ABC}{Meeds, Leenders, and Welling}
\firstpageno{1}

\author{
\name Edward Meeds \email tmeeds@gmail.com 
\AND
\name Robert Leenders \email robert.leenders@student.uva.nl 
\AND
Max Welling\email welling.max@gmail.com \\
              \addr Informatics Institute\\
              University of Amsterdam \\ 
              Amsterdam, Netherlands 
}
\begin{document}


                
\maketitle

%
%
\begin{abstract} 
  Approximate Bayesian computation (ABC) is a powerful and elegant framework for performing inference in simulation-based models.  However, due to the difficulty in scaling likelihood estimates, ABC remains useful for relatively low-dimensional problems. We introduce Hamiltonian ABC (HABC), a set of likelihood-free algorithms that apply recent advances in scaling Bayesian learning using Hamiltonian Monte Carlo (HMC) and stochastic gradients.     We find that a small number forward simulations can effectively approximate the ABC gradient, allowing Hamiltonian dynamics to efficiently traverse parameter spaces.  We also describe a new simple yet general approach of incorporating random seeds into the state of the Markov chain, further reducing the random walk behavior of HABC.  We demonstrate HABC on several typical ABC problems, and show that HABC samples comparably to regular Bayesian inference using true gradients on a high-dimensional problem from machine learning.
\end{abstract} 

\section{INTRODUCTION} \label{introduction}
In simulation-based science, models are defined by a simulator and its parameters.  These are called {\em likelihood-free} models because, in contrast to probabilistic models, their likelihoods are either intractable to compute or must be approximated by simulations.  To perform inference in likelihood-free models, a broad class of algorithms called Approximate Bayesian Computation \citep{beaumont2002approximate,marjoram2003markov,sisson2007sequential,sisson:2010,marin:2012,fan:2013} are employed.

At the core of every ABC algorithm is simulation.  To evaluate the quality of a parameter vector $\thetav$, a simulation is run using $\thetav$ as inputs and producing outputs $\x$.  If the pseudo-data $\x$ is ``close'' to observations $\y$, then $\thetav$ is kept as a sample from the approximate posterior.  Parameters $\thetav$ are then adjusted, depending upon the algorithm, to obtain the next sample.

In ABC, there is a fundamental trade-off between the computation required to obtain independent samples and the approximation to the true posterior.  If the parameter measuring closeness is too small, then samplers ``mix'' poorly; on the other hand, if it is too large, then the approximation is poor.  As the dimension of the parameters grows, the problem worsens, just as it does for general Bayesian inference with probabilistic models, but it is more acute for ABC due to its simulation requirement.  There is therefore a deep interest in improving the efficiency of ABC samplers (in terms of computation per independent sample).  In this paper we address this issue directly by using Hamiltonian dynamics to approximately sample from likelihood-free models with high-dimensional parameters.

Hamiltonian Monte Carlo (HMC) \citep{duane1987hybrid, neal2011mcmc} is perhaps the only Bayesian inference algorithm that scales to high-dimensional parameter spaces.  The core computation of HMC is the gradient of the log-likelihood.  Two problems arise if we consider HMC for ABC: one, how can the gradients be computed for high-dimensional likelihood-free models, and two, given a stochastic approximation to the gradient, can a valid HMC algorithm be derived?
 
 To answer the latter, we turn to recent developments in scaling Bayesian inference using HMC and stochastic gradients \citep{welling2011bayesian,chen2014stochastic,ding2014bayesian}.  We call these {\em stochastic gradient Hamiltonian dynamics} (SGHD) algorithms. SGHD are computationally efficient for two reasons.  First, they avoid computing the gradient of the log-likelihood over the entire data set, instead approximating it using small batches of data, i.e. computing stochastic gradients.  Second, they can maintain reasonable approximations to the Hamiltonian dynamics and therefore avoid a Metropolis-Hastings correction step involving the full data set.  Different strategies are employed to do this: small step-sizes combined with  Langevin dynamics \citep{welling2011bayesian}, using friction to prevent accumulation of errors in the Hamiltonian \citep{chen2014stochastic}, and using a thermostat to control the temperature of the Hamiltonian \citep{ding2014bayesian}.  Each of these strategies can be used by HABC.
 

In HABC, we use forward simulations to approximate the likelihood-free gradient. The key difference between SGHD methods and HABC is that the stochasticity of the gradient does not come from approximating the full data gradient with a mini-batch gradient, but by the stochasticity of the simulator.  It is therefore not the expense of the simulator (though this could very well be the case for many interesting simulation-based models -- see Section~\ref{sec:conclusion}) that requires an approximation to the gradient, but the likelihood-free nature of the problem.  

There are several difficulties in estimating gradients of likelihood-free models that we address with HABC.  The first is due to the form of the ABC log-likelihood.  As we show in Section~\ref{sec:abc}, using a conditional model for $\pi( \x | \thetav )$ provides an estimate of the ABC likelihood that is less sensitive to $\eps$ and therefore is more conducive to stochastic gradient computations.  The second difficulty is that for high-dimensional parameter spaces, computing the gradients naively (i.e. by finite differences \citep{kiefer1952stochastic}) can squash the gains brought by the Hamiltonian dynamics.  Fortunately, we can use existing stochastic approximation algorithms \citep{spall1992multivariate,spall2000adaptive} that can be used to compute unbiased estimators of the gradient with a small number of forward simulations that is {\em independent} of the parameter dimension.  The {\em stochastic perturbation stochastic approximation} (SPSA) \citep{spall1992multivariate} is described in Section~\ref{sec:habc}

A further innovation of this paper is the use of common random numbers (CRN) to improve the efficiency of the Hamiltonian dynamics.  The idea behind CRNs is to use the same set of random seeds for estimating a gradient by FD or SPSA, i.e. when simulating $\pi(\x| \theta + d\theta)$ and $\pi(\x| \theta - d\theta)$ use the same random seeds.  This was applied successfully to SPSA \citep{kleinman1999simulation} (and is analogous to using the same mini-batch in stochastic gradient methods).  We extend and simplify this approach by including the random seeds $\omega$ into the state of the Markov chain;  by keeping the random seeds fixed for several consecutive steps, the second order gradient stochasticity is greatly reduced.  We show that doing this produces a valid MCMC procedure.  This approach is not exclusive to HABC; our experiments show it also helps random-walk ABC-MCMC.

 
We briefly review ABC in Section~\ref{sec:abc}.  In Section~\ref{sec:scaling} we review three approaches to stochastic gradient inference using Hamiltonian dynamics: SGLD, SGHMC, and SGNHT.  We then introduce Hamiltonian ABC in Section~\ref{sec:habc}, where  we will show how to improve the stability of the gradient estimates by using CRNs and local density estimators of the simulator.  Extensions to high-dimensional parameter spaces are also discussed.  In Section~\ref{sec:demo} we show how HABC behaves on a simple one-dimensional problem, then in Section~\ref{sec:experiments} we compare HABC with ABC-MCMC for two problems: a low-dimensional model of chaotic population dynamics and a high-dimensional problem.

\section{APPROXIMATE BAYESIAN COMPUTATION}\label{sec:abc}
Consider the Bayesian inference task of either drawing samples from or learning an approximate model of the following (usually intractable) posterior distribution:
\begin{equation}
  \pi(\thetav | \y_1, \ldots, \y_N ) \propto \pi(\thetav) \pi( \y_1, \ldots, \y_N  | \thetav )
\end{equation}
where $\pi(\thetav)$ is a prior distribution over parameters $\thetav \in {\rm I\!R}^{D}$ and $\pi( \y_1, \ldots, \y_N  | \thetav )$ is the likelihood of $N$ data observations, where $\y_i \in {\rm I\!R}^J$.  In ABC, the vector of $J$ observations are typically informative statistics of the raw observations.  It can be shown that if the statistics used in the likelihood function are sufficient, then these algorithms sample correctly from an approximation to the true posterior \citep{marin:2012}.  
  The simulator is treated as generator of random pseudo-observations, i.e. $\x \simsim \pi( \x | \thetav )$ is a draw from the simulator.  Discrepancies between the simulator outputs $\x$ and the observations $\y$ are scaled by a closeness parameter $\eps$ and treated as likelihoods.  This is the equivalent to putting an $\eps$-kernel around the observations, and using a Monte Carlo estimate of the likelihood using $S$ draws of $\x$: 
\begin{equation}
  \pi_{\epsvec}( \y | \thetav ) =  \int \pi_{\epsvec}(\y | \x ) \pi( \x | \thetav ) d\x 
                           \approx  \frac{1}{S} \sum_{s=1}^S \pi_{\epsvec}(\y | \x^{(s)} ) \label{eq:abc_mc_approx}
\end{equation}

In ABC Markov chain Monte Carlo (MCMC) \citep{marjoram2003markov,Wilkinson2013,sisson:2010} the Metropolis-Hastings (MH) proposal distribution is composed of the product of the proposal for the parameters $\thetav$ and the proposal for the simulator outputs:
\begin{equation}
  q( \thetavp, \x^{(1)'}, \ldots, \x^{(S)'} | \thetav ) =  q( \thetavp | \thetav ) \prod_s \pi( \x^{(s)'} | \thetavp) \label{eq:pm-proposal}
\end{equation}
Using this form of the proposal distribution, and using the Monte Carlo approximation eq~\ref{eq:abc_mc_approx}, we arrive at the following Metropolis-Hastings accept-reject probability,
\begin{equation}
\alpha = \min \lp 1, \frac{\pi\lp\thetavp\rp \sum_{s=1}^S \pi_{\epsvec}(\y | \xps )  q( \thetav | \thetavp )}{\pi\lp\thetav\rp \sum_{s=1}^S \pi_{\epsvec}(\y | \xs ) q( \thetavp | \thetav )} \rp \label{eq:abc_mh_acceptance_with_s}
\end{equation}
If the simulations are part of the Markov chain, the algorithm corresponds to the pseudo-marginal (PM) sampler \citep{andrieu2009pseudo}, otherwise it is a marginal sampler \citep{marjoram2003markov,sisson:2010}.   For this paper we will be interested in the PM sampler because this is equivalent to having the random states that generated the simulation outputs in the state of the Markov chain, which we will use within a valid ABC sampling algorithm in Section~\ref{sec:habc}.
%

An alternative approach to computing the ABC likelihood is to estimate the parameters of a conditional model  $\pi(\x|\thetav)$, e.g. using kernel density estimate \citep{TurnerGenLik2014} or a Gaussian model \citep{wood2010statistical}.  While either approach should be adequate and both have their own limits and advantages, for this paper we will use a Gaussian model.  In ABC, using a conditional Gaussian model for  $\pi(\x|\thetav)$ is called a {\em synthetic likelihood} (SL) model \citep{wood2010statistical}.  For a SL log-likelihood model, we compute estimators of the first and second moments of $\pi(\x|\thetav)$.  The advantage is that for a Gaussian $\epsvec$-kernel, we can convolve the two densities   
\begin{eqnarray}
  \pi_{\epsvec}( \y | \thetav ) & = & \int \mathcal{N}( \y | \x, \epsvec^2 ) \mathcal{N}( \x | 
  \mu_{\thetav}, \sigma^2_{\thetav} ) d\x \\
                          & = & \mathcal{N}( \y | \mu_{\thetav}, \sigma^2_{\thetav} + \epsvec^2 )
\end{eqnarray}

Of particular concern to this paper is the behavior of the log-likelihoods for different values of $\epsvec$.  In the $\epsvec$-kernel case, the log-likelihood is very sensitive to small values of $\epsvec$:
\begin{eqnarray}
  \log \pi_{\epsvec}( \y | \thetav ) & = & \log \sum_s \mathcal{N}( \y | \x^{(s)}, \epsvec^2 ) \\
                          & = & \log \mathcal{N}( \y | \x^{(s)}, \epsvec^2 ) + \log\lp 1 + H \rp \\
                          & \approx & -\log \epsvec -\frac{1}{2 \epsvec^2} ( \y - \x^{(m)} )^2 
\end{eqnarray}
where $m$ is the simulation that is closest to $\y$, $H$ is a sum over terms close to $0$. We can see that the log-likelihood can be set arbitrarily small by decreasing $\epsvec$.  On the other hand, by using a model of the simulation at $\thetav$
\begin{eqnarray}
 \log \pi_{\epsvec}( \y | \thetav ) & \approx & -\frac{1}{2} \log (\sigma^2_{\thetav} + \epsvec^2 ) -\frac{( \y - \mu_{\thetav} )^2 }{2 (\sigma^2_{\thetav} + \epsvec^2 )} 
\end{eqnarray}
 For the SL model, $\epsvec$ acts as a smoothing term and can be set to small values with little change to the log-likelihood, as long as the SL estimators are fit appropriately.  This insensitivity to $\epsvec$ will be used in Section~\ref{sec:habc} for estimating gradients of the ABC likelihood.  Before describing HABC in full detail however, we now explain how scaling Hamiltonian dynamics in Bayesian learning can be accomplished using stochastic gradients from batched data.
 
\section{SCALING BAYESIAN INFERENCE USING HAMILTONIAN DYNAMICS} \label{sec:scaling}
Scaling Bayesian inference algorithms to massive datasets is necessary for their continuing relevance in the so-called {\em big data} era.  We now review the role stochastic gradient methods combined with Hamiltonian dynamics have played in recent advances in scaling Bayesian inference.   Most importantly, these methods have combined the ability of HMC to explore high-dimensional parameter spaces with the computational efficiency of using stochastic gradients based on small mini-batches of the full dataset.  After an overview of HMC, we will briefly describe stochastic gradient Hamiltonian dynamics (SGHDs), starting with using  Langevin dynamics \citep{welling2011bayesian}, then HMC with friction \citep{chen2014stochastic}, and finally HMC with thermostats \citep{ding2014bayesian}.  We will then make the connection between SGHDs and HABC in Section~\ref{sec:habc}.

\subsection{Hamiltonian Monte Carlo}\label{sec:hmc}

Hamiltonian dynamics are often necessary to adequately explore the target distribution of high-dimensional parameter spaces.  By proposing parameters that are far from the current location and yet have high acceptance probability, Hamiltonian Monte Carlo \citep{duane1987hybrid, neal2011mcmc}  can efficiently avoid random walk behavior that can render proposals in high-dimensions painfully slow to mix.

HMC simulates the trajectory of a particle along a frictionless surface, using random initial momentum $\rhov$ and position $\thetav$.  The Hamiltonian function computes the energy of the system and the dynamics govern how the momentum and position change over time.  The continuous Hamiltonian dynamics can be simulated by discretizing time into small steps $\eta$.  If $\eta$ is small, the value of $\thetav$ at the end of a simulation can be used as proposals within the Metropolis-Hastings algorithm.  Hamiltonian dynamics should propose $\thetav$ that are always accepted, but errors due to discretization may require a  Metropolis-Hastings correction.  It is this correction step that SGHD algorithms want to avoid as it requires computing the log-likelihood over the full data set.

More formally, the Hamiltonian $H\lp\thetav, \rhov \rp = U(\thetav) + K(\rhov)$ is a function of the current potential energy $U(\thetav)$ and kinetic energy $K(\rhov) = \rhov^T M^{-1}\rhov/2$ ($M$ is a diagonal matrix of masses which for presentation are set to $1$).  The potential energy is defined by the negative log joint density of the data and prior:
\begin{equation}
  U(\thetav) = - \log \pi(\thetav) - \sum_{i=1}^N \log \pi(\yi | \thetav )
\end{equation}
The Hamiltonian dynamics follow 
\begin{equation}
  d\thetav = \rhov dt ~~~~~~~~~~~~ d\rhov = -\nabla U(\thetav) dt
\end{equation}
in simulation $dt = \eta$. 

\subsection{Stochastic Gradient Hamiltonian Dynamics}\label{se:sghd}
If the log-likelihood over the full data set is replaced with a mini-batch estimate, as is done for the following {\em stochastic gradient Hamiltonian dynamics} (SGHDs) algorithms, then the error in simulating the Hamiltonian dynamics comes not only from the discretization, but from the variance of the stochastic gradient.  As long as this error is controlled, either by using small steps $\eta$ (SGLD), or adding friction terms $B$ (SGHMC), or using a thermostat $\xi$ (SGNHT), the expensive MH correction step can be avoided and values of $\thetav$ from the Hamiltonian dynamics can be used as approximate samples from the posterior.

SGHDs replace the full potential energy and its gradient with a mini-batch approximation:
\begin{eqnarray}
  \hat{U}(\thetav)       & = & - \log \pi(\thetav) - \frac{N}{n} \sum_{i=h_1}^{h_n} \log \pi(\yi | \thetav ) \\
  \nabla\hat{U}(\thetav) & = & - \nabla \log \pi(\thetav) - \frac{N}{n} \sum_{i=h_1}^{h_n} \nabla \log \pi(\yi | \thetav ) 
\end{eqnarray} 
where $n$ is the mini-batch size, and $h_i$ are indices chosen randomly without replacement from $[1,N]$ (i.e. it defined a random mini-batch).  
 
{\bf Stochastic gradient Langevin dynamics} \citep{welling2011bayesian} performs one full leap-frog step of HMC.   Starting with a half step for the momentum, the update for $\thetav$ is 
\begin{eqnarray}
  \rhov_t & \sim & \mathcal{N}(0,\eye_p) \\
  \rhov_{t+\frac{1}{2}} & = & \rhov_t - \eta \nabla \hat{U}(\thetav_t) /2 \\
  \thetav_{t+1} & = & \thetav_t + \eta \rhov_{t+\frac{1}{2}}
\end{eqnarray}
It is not necessary to include $\rhov$ in the updates since there is only one step:
\begin{eqnarray}
  \thetav_{t+1} & = & \thetav_t + \eta \mathcal{N}(0,\eye_p) - \eta^2 \nabla \hat{U}(\thetav_t) /2 
\end{eqnarray}
One of the potential drawbacks of SGLD is that the momentum term is {\em refreshed} for every update of the parameters, and since this means the parameter update only uses the current gradient approximation, it limits the benefits of using Hamiltonian dynamics.  On the other hand, this also prevents SGLD from accumulating errors in the Hamiltonian dynamics.

{\bf Stochastic Gradient HMC} (SGHMC) \citep{chen2014stochastic} avoids $\rhov$ refreshment altogether.  By applying HMC directly using the stochastic approximation $\hat{U}$ and $\nabla \hat{U}$, which the authors call {\em naive SGHMC}, the variance of the gradient will introduce errors that left unaddressed will result in sampling from the incorrect target distribution.  Under the assumption that $\nabla \hat{U}(\thetav) = \nabla U(\thetav) + \mathcal{N}\lp \zerov, \varv_{\thetav}\rp$, where $\varv_{\thetav}$ is the covariance of the gradient approximation, and  updates $\rhov_{t+1} = \rhov_{t} + \Delta \rhov_{t}$ and $\thetav_{t+1} = \thetav_t + \eta \rhov_{t+1}$, the change in momenta $\Delta \rhov$ from one full step is
 \begin{equation}
   - \eta \lp \nabla U(\thetav) + \mathcal{N}\lp \zerov, \varv_{\thetav}\rp\rp  =  - \eta \nabla U(\thetav) + \mathcal{N}\lp \zerov, \eta^2 \varv_{\thetav}\rp  
 \end{equation}
By adding a friction term $\Bv$ to $\Delta \rhov$ proportional to $\varv_{\thetav}$, the correction step can be avoided
 \begin{eqnarray}
   \Delta \rhov & = & - \eta \Bv \rhov_t - \eta \nabla U(\thetav_t) + \mathcal{N}\lp \zerov, 2 \eta \Bv\rp 
 \end{eqnarray}
 where  $\Bv=\frac{1}{2}\eta \varv_{\thetav_t}$.   In practice, since we can only estimate $\Bv$ by some $\hat{\Bv}$ and can only compute $\hat{U}$, a user defined friction term $\Cv$ is used (with $\Cv - \hat{\Bv}$ is semi-positive definite).  Thus the updates used for $\Delta \rhov$ for SGHMC:
 \begin{equation}
   - \eta \Cv \rhov_t - \eta \nabla \hat{U}(\thetav_t) + \mathcal{N}\lp \zerov, 2 \eta (\Cv-\hat{\Bv}) \rp 
 \end{equation}
 In our experiments we compute an online estimate $\hat{\varv}$ and set $\Cv = c \eye_p + \hat{\varv}$.  
 
 {\bf Stochastic Gradient thermostats} (SGNHT) \citep{ding2014bayesian} addresses the difficulty of estimating $\hat{\Bv}$ by introducing a scalar variable $\xi$ who's addition to the Hamiltonian dynamics maintains the temperature of the system constant, i.e. it acts as a (Nos\'{e}-Hoover) thermostat \citep{leimkuhler2009metropolis}.  The update equations remain simple: initialize $\xi = \Cv$ (or $c$), then for $t=1\ldots$
 \begin{eqnarray}
   \rhov_{t+1} & = & \rhov_t - \eta \xi_t \rhov_t - \eta \nabla \hat{U}(\thetav_t) + \mathcal{N}\lp \zerov, 2 \eta_t \Cv \rp \\
   \thetav_{t+1} & = & \thetav_t + \eta \rhov_{t+1} \\
   \xi_{t+1} & = & \xi_t + \eta \lp  \rhov_{t+1}^T \rhov_{t+1} / D - 1 \rp
 \end{eqnarray}
 In summary, the hyperparameters required for these algorithms are $\eta$ and $\Cv$ (for SGHMC and SGNHT only), and in practice, some way of estimating $\hat{\varv}$ for SGHMC.  
  
\section{HAMILTONIAN ABC}\label{sec:habc}
The general approach of applying Hamiltonian dynamics to ABC requires choosing one of the SGHD algorithms and then plugging in the ABC gradient approximation $\nabla \hat{U}(\thetav)$.  With this in mind we leave the details of the Hamiltonian updates to previous work \citep{welling2011bayesian,chen2014stochastic,ding2014bayesian} and focus on the details of how stochastic gradients are computed in the likelihood-free setting.

\subsection{Deterministic Representations of Simulations}
Implicit in each simulation run $\x \simsim \pi(\x | \thetav)$ is a sequence if internally generated random numbers that are used to produce random draws from $\pi(\x | \thetav)$.  These random numbers are important to HABC because we wish to control the stochasticity of the simulator when computing its gradient.  Furthermore, we will control the random numbers over multiple time steps.  Instead of keeping track of random numbers, we can equivalently keep a vector of $S$ random seeds $\omegav$.  This allows HABC to treat the simulation function $\pi(\x|\thetav)$ as a blackbox, outside of which we can control the random number generator (RNG), and represent $\x^{(s)}$ as the output of a deterministic function; i.e.  $\x^{(s)} = f( \thetav, \omega_s)$ instead of $\x^{(s)} \simsim \pi(\x | \thetav)$.   We include $\omegav$ as part of the state of our Markov chain.

\subsection{Kernel-$\epsvec$ versus Synthetic-likelihood -based Gradients}\label{sec:habc-grads}
In Section~\ref{sec:abc} we showed that the synthetic-likelihood representation of $\lleps(\thetav)$ is less sensitive to small choices of $\epsvec$.  This is particularly important to HABC as our gradient approximations are proportional to differences in $\lleps(\thetav)$; if the variance of the stochastic gradients is too high, then we must choose a very small step-size $\eta$, eliminating the usefulness of HMC for ABC.  Under the deterministic representation of $\x^{(s)}$, we can write the loglikelihood as
\begin{eqnarray}
 \lleps(\thetav) & \propto & \log \sum_s \mathcal{N}( \y | f( \thetav, \omega_s), \epsvec^2 )\\
                 & \approx & -\log \epsvec -\frac{1}{2 \epsvec^2} ( \y - f( \thetav, \omega_m) )^2 
\end{eqnarray}
In the second line we have assumed $\epsvec$ is very small and $m$ is the index of the random seed producing the closest simulation to $\y$.  For a finite difference approximation, $\partial \lleps(\thetav)/\partial\thetav$ is
\begin{equation}
 \frac{1}{4\dtheta \epsvec^2} \lp ( \y - f( \thetav-\dtheta, \omega_m^-) )^2 -( \y - f( \thetav + \dtheta, \omega_m^+) )^2 \rp
\end{equation}

\begin{figure}[t]
\begin{center}
\includegraphics[width=0.5\columnwidth]{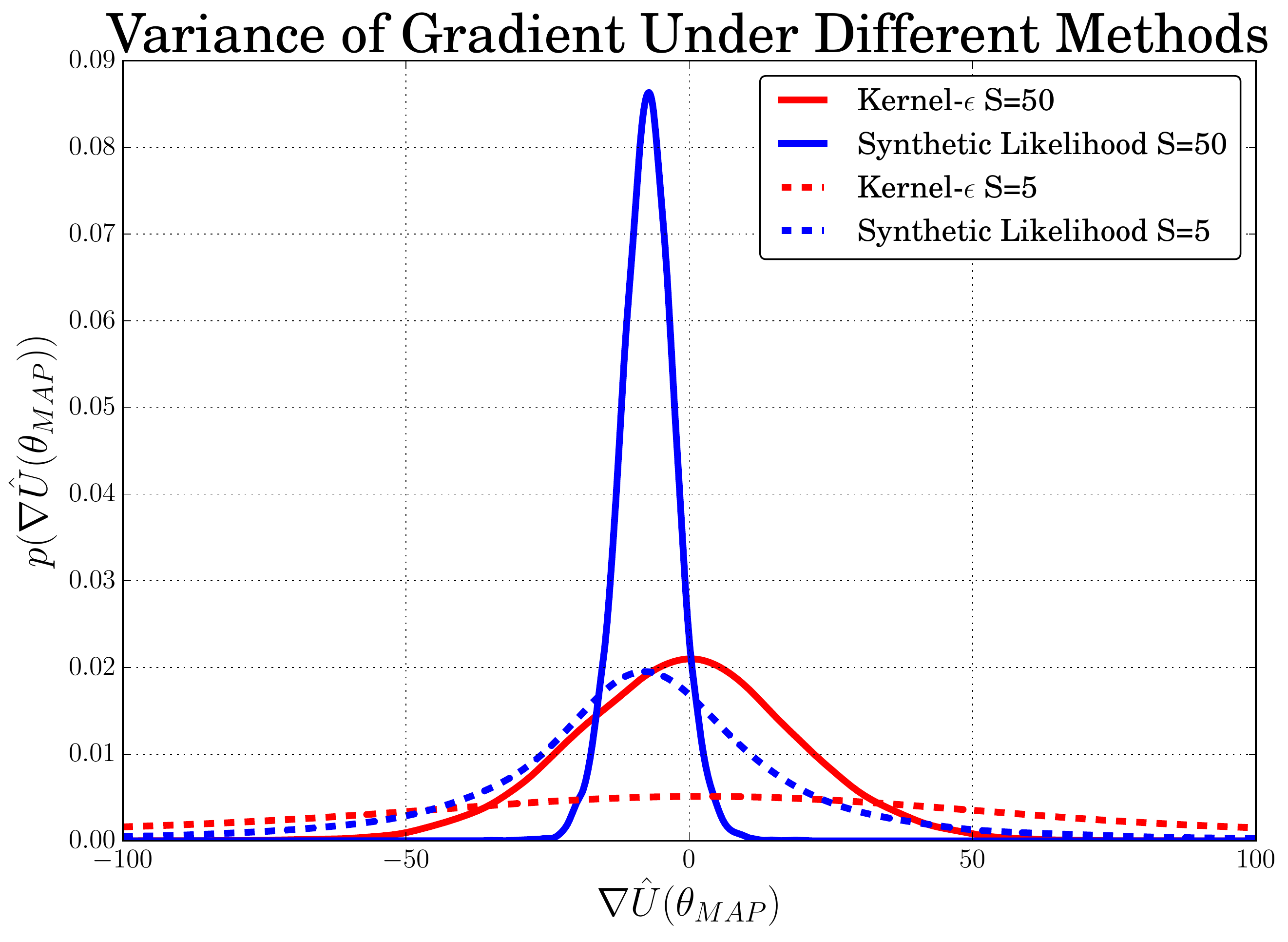}
\caption{\small{Variance of gradient estimation using kernel-$\eps$ and SL for different values of $S\in\{5,50\}$ and fixed $\eps = 0.37$ (the same used in the other results).  
When $S=5$, the empirical estimates of  $\nabla \hat{U}(\thetav_{\text{MAP}})$ are $-12 \pm 147$ (kernel-$\eps$) and $-9.3 \pm 43$ (SL).  When $S=50$ they are $-0.80 \pm 19$ (kernel-$\eps$) and $-7.3 \pm 4.9$ (SL).  Note the large discrepancy in variance.  Note the limit of $S \rightarrow \infty$,   $\nabla \hat{U}(\thetav_{\text{MAP}}) = -7.8$.  The bias if SL gradients is due to its Gaussian approximation (smoothed by $\eps$) of $\pi(\x|\thetav)$, which is a heavy-tailed Gamma distribution (the sum of $N$ exponentials).}}
\label{fig:exp-varg}
\end{center}
\end{figure}

On the other hand, the synthetic-likelihood is stable; using a deterministic representation, we have 
\begin{equation}
\mu_{\thetav} = \frac{1}{S} \sum_s f( \thetav, \omega_s ) ~~~~~~~~~~~~~ \sigma^s_{\thetav} = \frac{1}{S-1} \sum_s ( \mu_{\thetav} - f( \thetav, \omega_s ) )^2
\end{equation}
the gradients (for a 1-dim problem) use $\epsvec$ as a smoothness prior in $\partial \lleps(\thetav)/\partial\thetav$:
\begin{equation}
  -\frac{1}{2}\log\lp\frac{\sigma_{\theta^+}^2 + \epsvec^2}{\sigma_{\theta^-}^2 + \epsvec^2}\rp -\frac{( \y - \mu_{\thetav^+} )^2 }{2 (\sigma^2_{\thetav^+} + \epsvec^2 )} +\frac{( \y - \mu_{\thetav^-} )^2 }{2 (\sigma^2_{\thetav^-} + \epsvec^2 )} 
\end{equation} 
In Figure~\ref{fig:exp-varg}, as part of our demonstration of HABC, we compare the gradient approximations around the true $\thetav_{\text{MAP}}$ using SL versus kernel-$\epsvec$ Figure~\ref{fig:exp-varg} for a simple problem.  We find that although, for this particular problem, SL has a small bias due to its Gaussian assumption, it has much smaller variance, an important property for HABC.

\subsection{From Finite Differences to Simultaneous Perturbations}
\begin{algorithm}[t]
	\caption{$\nabla U$ FDSA-ABC}\label{algo:fdsa}
	\begin{algorithmic}
		\STATE {\bfseries inputs:} $\thetav, \dtheta, f, \omegav, \lleps, \pi$ 
		\STATE $\gradvest \leftarrow \zerov$
			\FOR{$r = 1 : |\thetav|$}
        \STATE $\Deltav \leftarrow \zerov$ 
        \STATE $\Deltavr \leftarrow 1$
        \FOR{$s = 1 : |\omegav|$}
          \STATE $\x^{(s)}_{+} \leftarrow f\lp \thetav + \dtheta \Deltav, \omega_s \rp$
          \STATE $\x^{(s)}_{-} \leftarrow f\lp \thetav - \dtheta \Deltav, \omega_s \rp$
        \ENDFOR
        \STATE $\gradvest_r \leftarrow \lleps( \{\x^{(s)}_{+}\}) - \lleps( \{\x^{(s)}_{-}\})$
			\ENDFOR
    \STATE $\gradvest \leftarrow \gradvest/ 2\dtheta + \nabla \log \pi(\thetav)$
		\STATE \textbf{return} $-\gradvest$
	\end{algorithmic}
\end{algorithm}
\begin{algorithm}[t]
	\caption{$\nabla U$ SPSA-ABC}\label{algo:spsa}
	\begin{algorithmic}
		\STATE {\bfseries inputs:} $\thetav, \dtheta, f, \omegav, \lleps, \pi, R$ 
		\STATE $\gradvest \leftarrow \zerov$
			\FOR{$r = 1 : R$}
        \STATE $\Deltav \sim 2 \cdot \text{Bernouilli}\lp 1/2, |\thetav| \rp$ - 1
        \FOR{$s = 1 : |\omegav|$}
          \STATE $\x^{(s)}_{+} \leftarrow f\lp \thetav + \dtheta \Deltav, \omega_s \rp$
          \STATE $\x^{(s)}_{-} \leftarrow f\lp \thetav - \dtheta \Deltav, \omega_s \rp$
        \ENDFOR
        \STATE $\gradvest \leftarrow \gradvest + \lp \lleps( \{\x^{(s)}_{+}\}) - \lleps( \{\x^{(s)}_{-}\}) \rp \cdot \Deltav^{-1}$
			\ENDFOR
    \STATE $\gradvest \leftarrow \gradvest/(2\dtheta R) + \nabla \log \pi(\thetav)$
		\STATE \textbf{return} $-\gradvest$
	\end{algorithmic}
\end{algorithm}
Algorithm~\ref{algo:fdsa} shows the {\em finite difference stochastic approximation} (FDSA) \citep{kiefer1952stochastic} to $\nabla U(\thetav)$ as a function of random seeds $\omegav$.  Note we have deliberately shown the deterministic simulations ($f$) outside of $\lleps$ to emphasize its dependence on $\x$.  The number of simulations required for FDSA is $2 S D$, which may be acceptable for some small ABC problems.  Our goal is to scale ABC to high-dimensions and for that we need an alternative stochastic approximation of $\nabla U(\thetav)$.

In the gradient-free setting, Spall \citep{spall1992multivariate, spall2000adaptive} provides a stochastic approximate to the true gradient using only $2$ forward simulations for any dimension $D$ (though the approximation can be improved by averaging $R$ estimates).  Spall's {\em simultaneous perturbation stochastic approximation} (SPSA) algorithm works as follows. Let $L$ be the gradient-free function we wish to optimize.  Each approximation randomly generates a {\em perturbation mask} (our name) $\Deltav$ of dimension $D=|\thetav|$ where entry $\Deltavd \sim 2 \text{Bernouilli}(1/2) - 1$.  Then $L$ is evaluated at $\thetav+\dtheta \Deltav$ and $\thetav-\dtheta \Deltav$, giving the gradient approximation $\gradvest(\thetav) \approx \partial L(\thetav) / \partial \thetav$:
\begin{equation}
  \gradvest(\thetav) = \frac{L\lp\thetav+\dtheta \Deltav\rp - L\lp\thetav-\dtheta \Deltav\rp}{2 \dtheta} \begin{bmatrix} 
                                   1/\Delta_1 \\
                                   1/\Delta_2 \\
                                   \vdots \\
                                   1/\Delta_D \\
                                \end{bmatrix}
\end{equation}
If we let $\gradvest_r(\thetav)$ be the estimate using perturbation mask $\Deltav_r$, the estimate $\gradvest(\thetav)$ can be improved by averaging $\gradvest(\thetav) = 1/R \sum_r \gradvest_r(\thetav)$.
Algorithm~\ref{algo:spsa} shows SPSA to estimate $\nabla U(\thetav)$.  The number of simulations required for SPSA is $2 S R$, where $R \geq 1$.

Variations of SPSA include {\em one-sided} SPSA \citep{spall2000adaptive} (we use what Spall calls 2SPSA) and an algorithm for estimating the Hessian based on the same principle as SPSA \citep{spall2005monte}.   The one-sided version is attractive computationally, but for HABC, the updates for $\thetav$ require simulating two-sides anyway (once at $\thetav$, after an step, and once for the one-sided gradient), so using 2SPSA makes more sense.  SPSA has also been used within a procedure for maximum-likelihood estimation for hidden Markov models using ABC \citep{Ehrlich2013}.

\begin{figure}[t]
\vskip 0.2in
\begin{center}
\includegraphics[width=0.5\columnwidth]{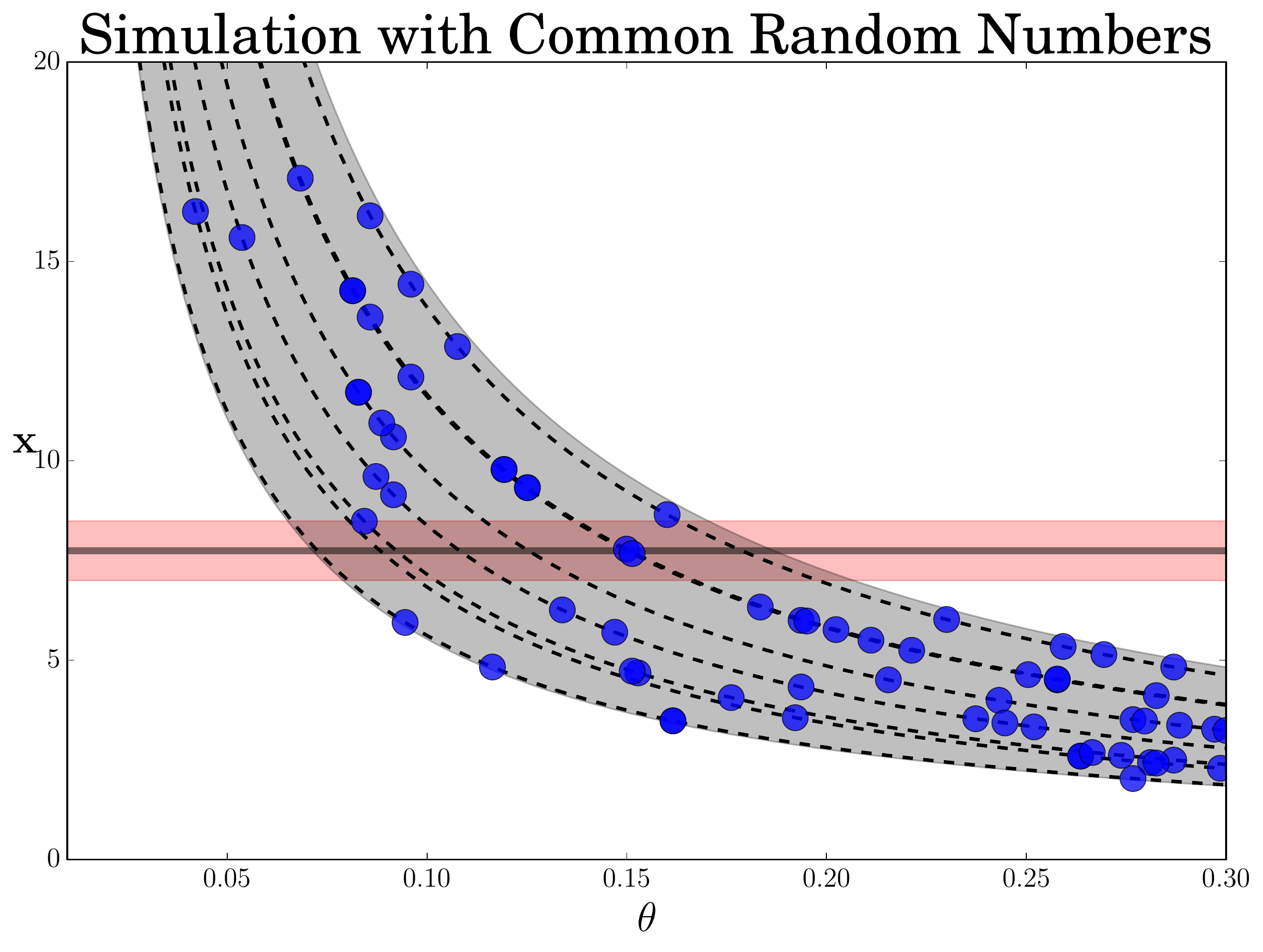}
\caption{\small{A view of a simulator in terms of common random numbers.  The horizontal line represents $\y$ and red shading $\pm 2\eps$.  The shaded curved region represents $2\sigma$ of $\pi(\x|\thetav)$.  The dashed lines are $f(\thetav, \omega_s)$ for several values of $\omega$.  The blue circles are potential random samples from $\pi(\x|\thetav)$.  For a fixed value $\omega_s$, the simulator produces deterministic outputs that change smoothly, even though the simulator itself is quite noisy.}}
\label{fig:exp-crns}
\end{center}
\end{figure} 

\subsection{Common and Sticky Random Numbers}
The usefulness of applying common random numbers (CRNS) in SPSA has been previously demonstrated \citep{kleinman1999simulation}.  In that work, the same random numbers are used to simulate both sides of the optimization function within the SPSA gradient.  This makes sense intuitively, as we would generally assume that the expected simulation function varies smoothly in $d \thetav$;  by using CRNs, this smoothness is easily exploited (see Figure~\ref{fig:exp-crns}).  If we were to apply SPSA to Bayesian learning, then using CRNs in the gradient step would be analogous to using the same mini-batch for both sides of the computation.  

In addition to using CRNs in simulations for each gradient computation, we have found that using {\em persistent} random {\em seeds} helps  HABC explore the parameter landscape more easily for some algorithms and problems.  Intuitively, for a gradient-based sampling algorithm, it means a particle can slide along a smooth Hamiltonian landscape because    the additive noise is suppressed.  This is very similar to using dependent random streams to drive MCMC \citep{Murray2012,Neal2012}, the main difference we believe is that we are using the Hamiltonian dynamics to drive proposals for $\thetav$ and using persistent seeds $\omegav$ to suppress simulation noise.

  
Using random seeds (versus, say, a set of random numbers) allows us to treat the simulator as a black-box, setting the random seed of its RNG without knowing the internal mechanisms it uses to generate random numbers.  In light of our arguments above, we propose including persistent random seeds $\omegav$ in the state of our Markov chain.    We will now describe a simple  Metropolis-Hastings transition operator that randomly proposes {\em flipping} each seed $\omega_s$ at time $t$ with some probability $\gamma$.  

This Metropolis-Hastings transition conditions of the current parameter location $\thetav$ and proposes changing a single random seed $\omega$ (it easily generalizes to $S$ seeds).  The procedure is as follows: 1) propose a new seed $\omega^{'} \sim q(\omega^{'} | \omega) = \pi(\omega)$ (independent of the current seed and from its uniform prior); 2) simulate {\em deterministically} $\x^{'} = f( \thetav, \omega^{'})$; 3) compute the acceptance ratio (which reduces to the ratio of $\pi(\y | \x^{'})/\pi(\y | \x)$).   It is straightforward to show that this leaves the target distribution invariant.  The probability of the proposal is $q(x^{'}, \omega^{'} | \thetav, \omega) = \pi(\omega^{'})\delta( \x^{'} - f( \thetav, \omega^{'}))$, where $\delta(a)$ is a delta function at $a=0$.  Because the $q$ has this form, acceptance ratio simplifies:
\begin{equation}
   \frac{\pi_{\epsvec}(\y | \x^{'})\pi(\omega^{'}) \pi(\x^{'} | \thetav, \omega_{'}) }{\pi_{\epsvec}(\y|\x)\pi(\omega)\pi(\x | \thetav, \omega)} \frac{\pi(\omega)\delta( \x - f( \thetav, \omega))}{\pi(\omega^{'})\delta( \x^{'} - f( \thetav, \omega^{'}))} =  \frac{\pi_{\epsvec}(\y | \x^{'})}{\pi_{\epsvec}(\y | \x)}
\end{equation}
In pseudo-marginal ABC-MCMC one could propose $q(\x^{'(s)} | \thetav)$ (fixing $\thetav$) and still sample correctly from the distribution of simulations with high likelihood at $\thetav$.  What we propose is slightly different. By instead keeping the random seeds fixed, we can sample $\thetav$ using HABC and use $\omegav$ as CRNs within the gradient computation step and suppress gradient noise over time.  In this way, random seeds carry over the same additive noise from one step to the next.
\begin{figure}[t]
\begin{center}
\includegraphics[width=0.22\columnwidth]{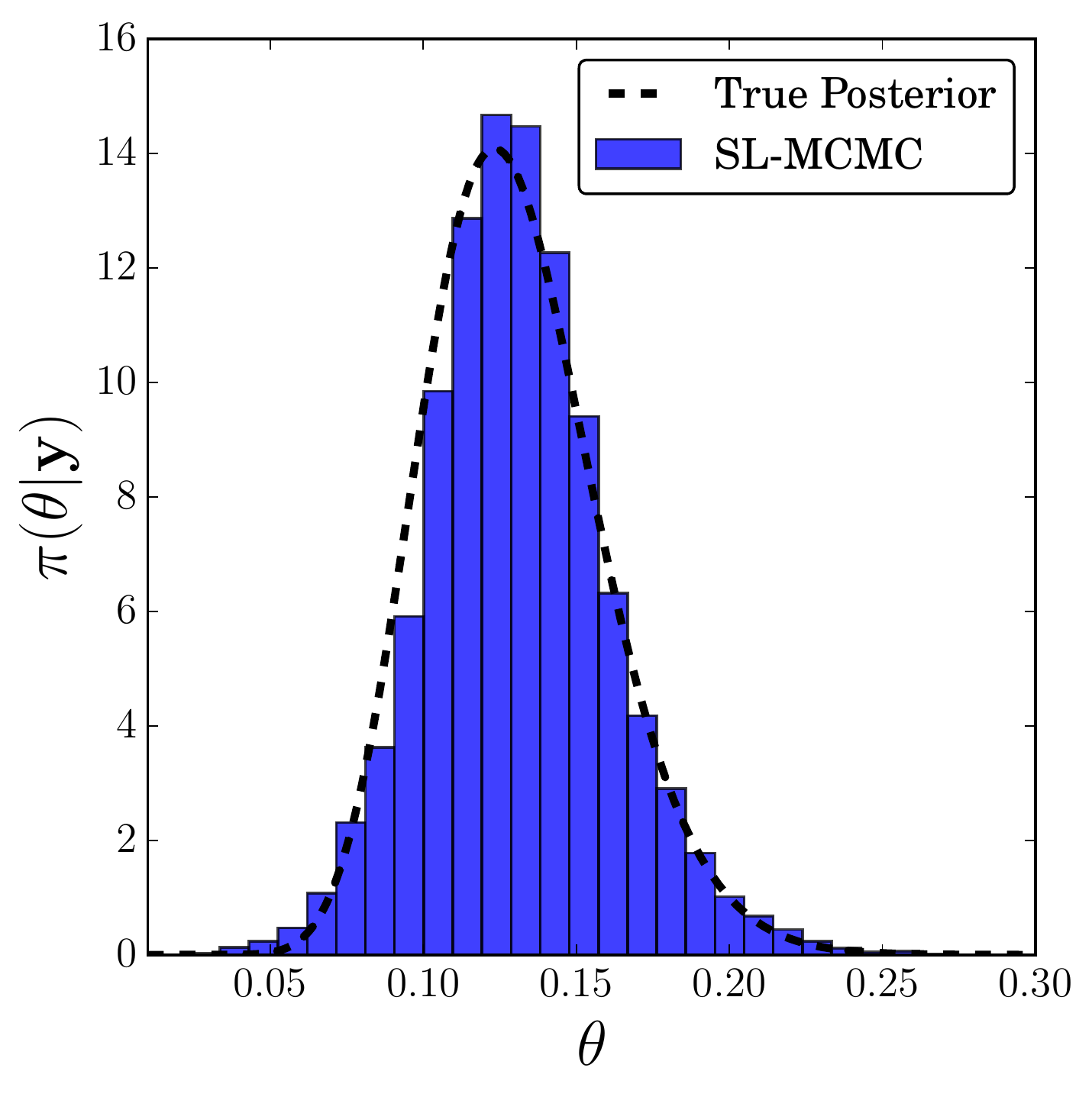}
\includegraphics[width=0.22\columnwidth]{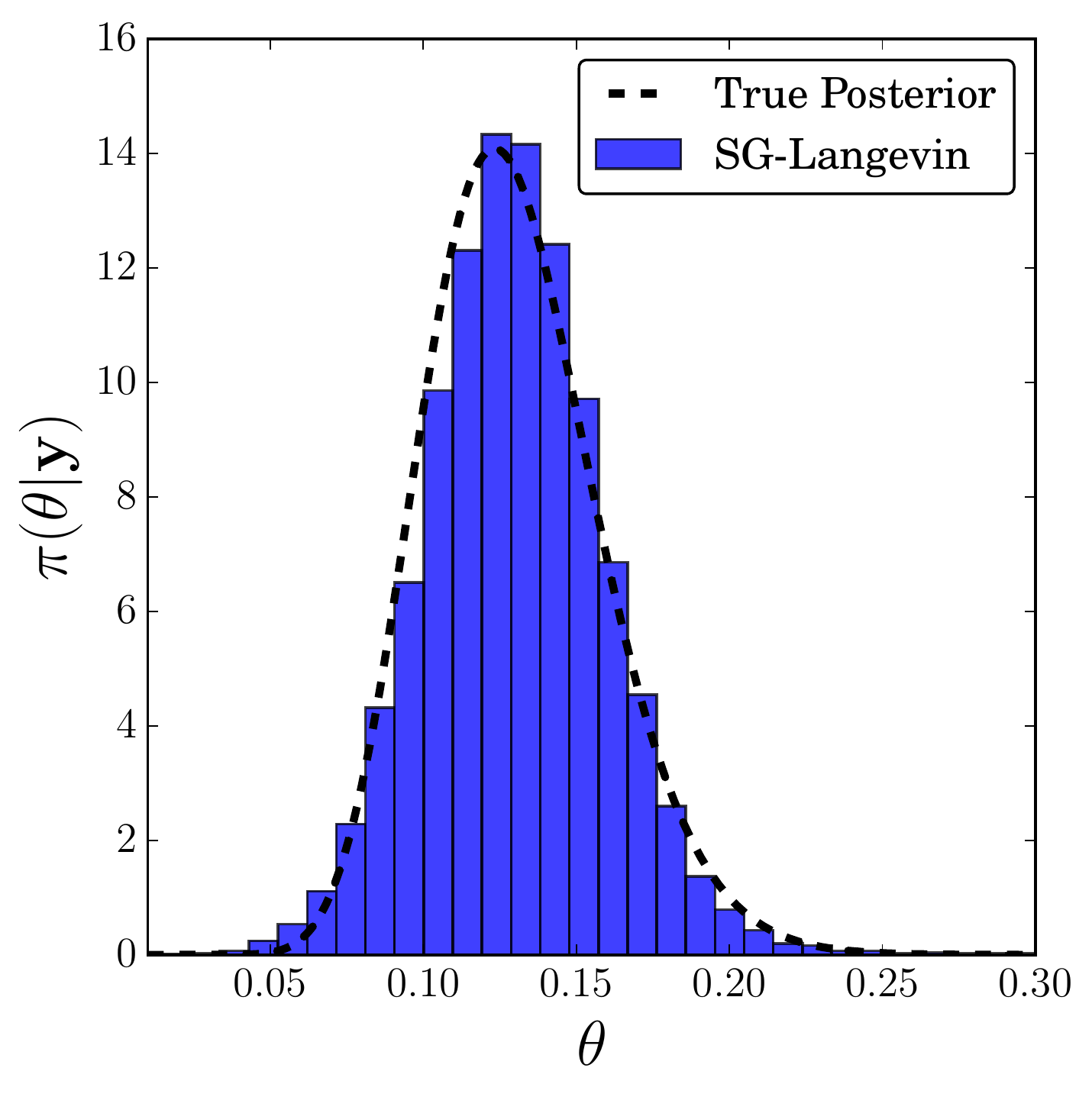}
\includegraphics[width=0.22\columnwidth]{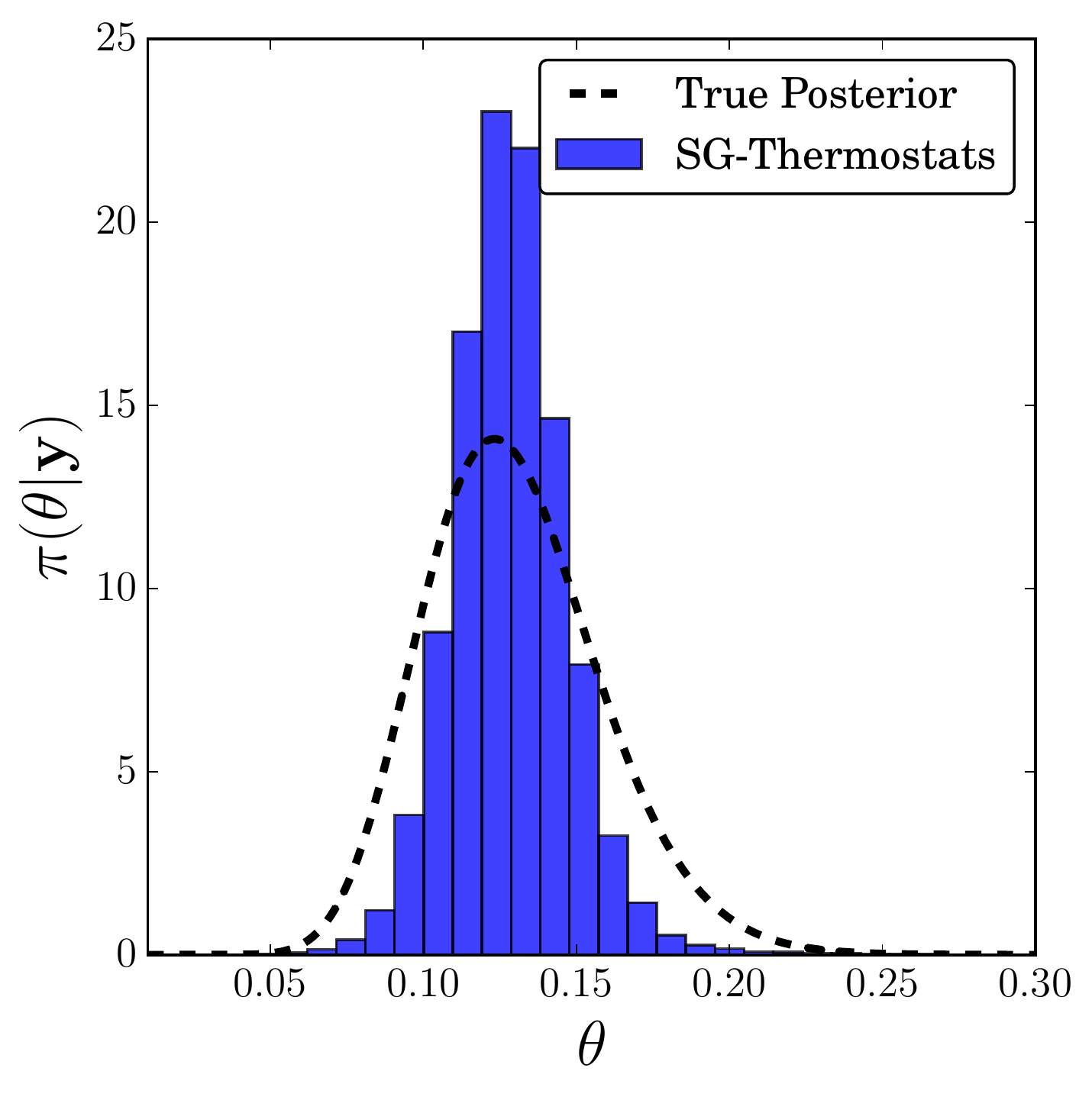}

\includegraphics[width=0.22\columnwidth]{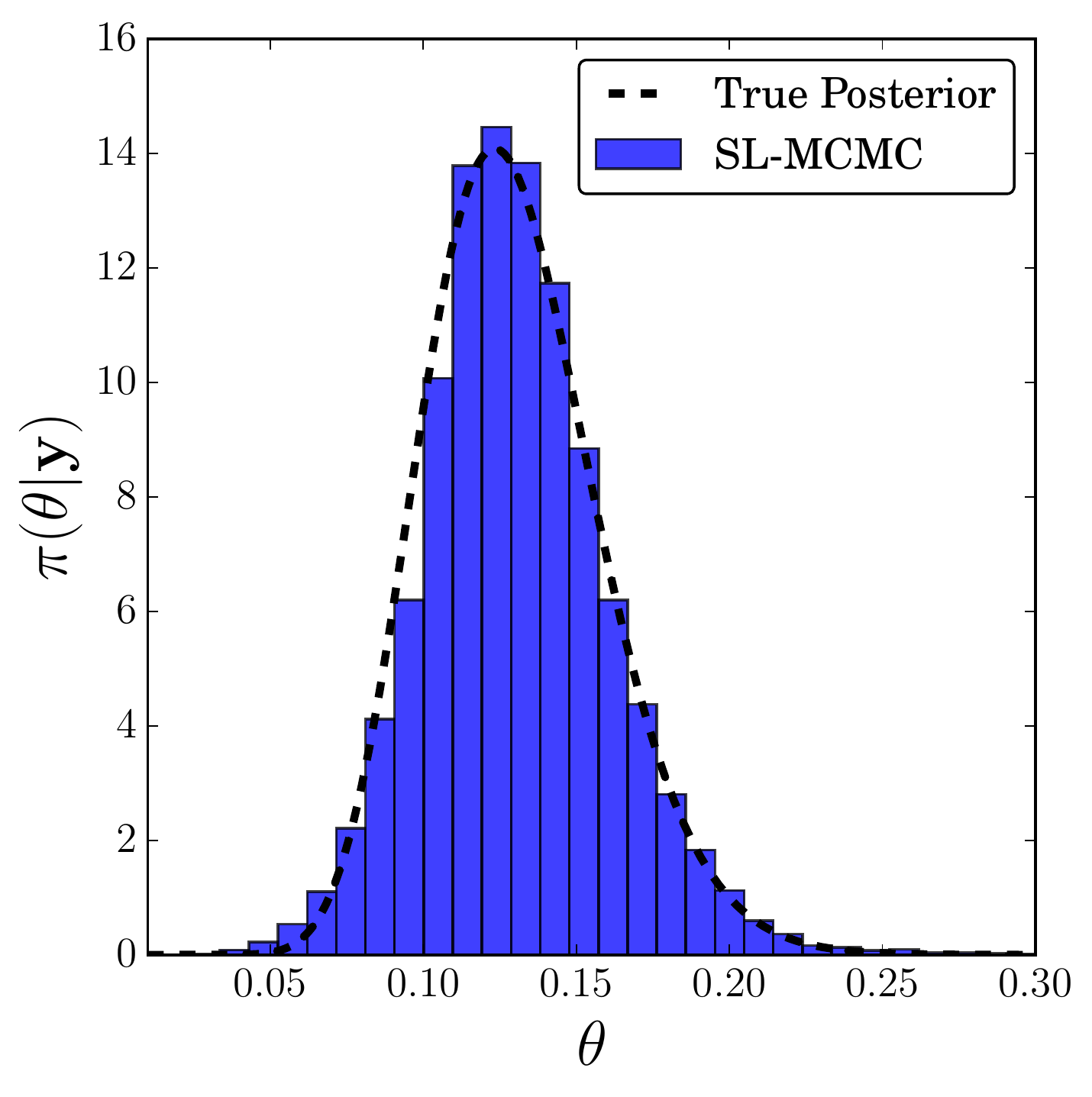}
\includegraphics[width=0.22\columnwidth]{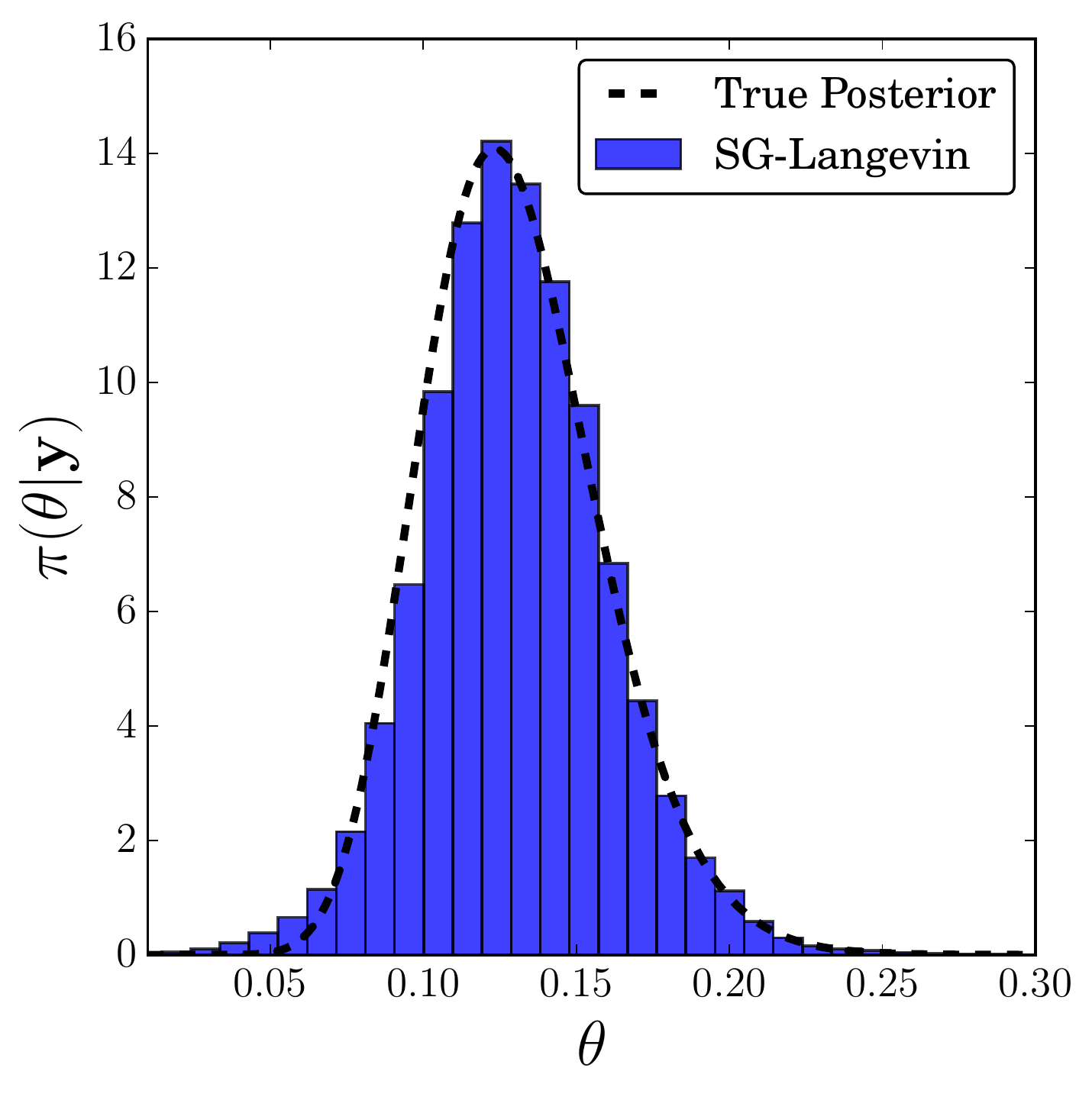}
\includegraphics[width=0.22\columnwidth]{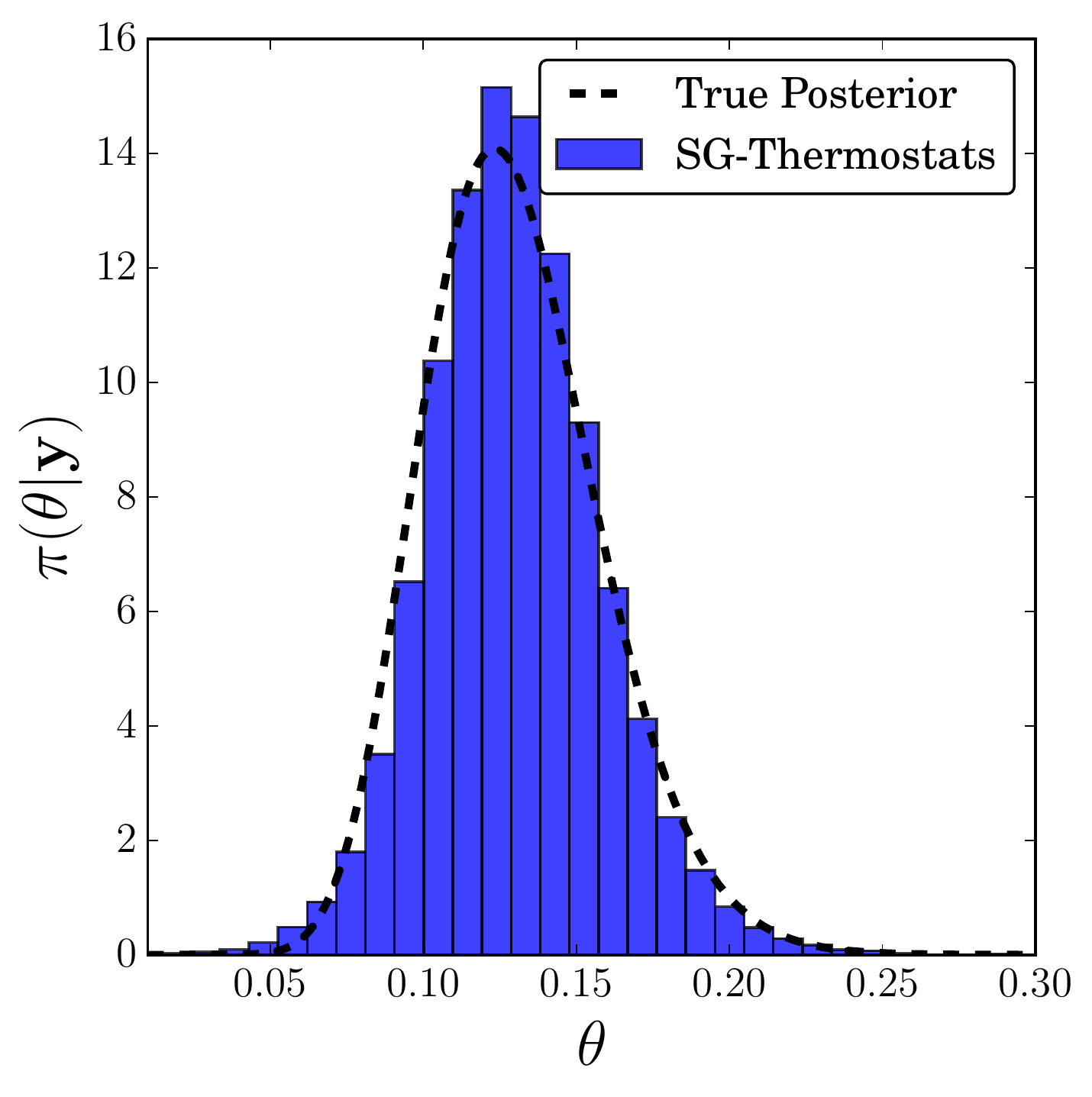}
\vspace{-0.1in}
\caption{\small{Posterior distributions for the demonstration problem.  {\bf Top row:} No persistent seeds.  {\bf Bottom row:} Persistent seeds with $\gamma=0.1$.  Histograms of the posterior estimates are overlaid with the true posterior (dashed line).  All algorithms (except for SG-Thermostats for non-persistent $\omegav$) give roughly the same posterior estimate.  By adding persistent $\omegav$ SG-Thermostats achieved similar posteriors to the other algorithms.}}
\label{fig:exp-posteriors}
\end{center}
\end{figure} 
%

\section{Demonstration}\label{sec:demo}
We use a simple $D=1$ problem to demonstrate HABC.  Let $y= \frac{1}{N} \sum_{i} e_i$, where $e_i \sim \text{Exp}(1/\thetastar)$; $\thetastar = 0.15$, $N=20$, and $y=7.74$ in our concrete example.  Assuming $\pi(\theta ) = \text{Gamma}(\alpha, \beta)$, the true posterior is a gamma distribution with shape $\alpha+N$ and rate $\beta + N y$.  Our simulator therefore generates the average of $N$ exponential random variates with rate $\lambda = 1/\theta$.  Data $x \simsim \pi(x|\theta)$ are shown in Figure~\ref{fig:exp-crns}.  We have explicitly shown the smoothness of the simulator by generating data along trajectories of fixed seeds $\omega_s$; i.e. for several $\omega_s$ we vary $\theta$ (dashed lines are function $f(\theta, \omega_s)$) and randomly reveal simulation data (blue circles).  The horizontal line with shading indicates $y \pm 2 \eps$, where $\eps = 0.37$ is used throughout the demonstration.

\subsection{Bias and Variance of $\nabla \hat{U}(\thetav)$}

To test our assumption that the synthetic-likelihood model is better suited for HABC, we ran FDSA at the true $\theta_{\text{MAP}}$.  Using $S=5$ and $S=50$ and fixing $\eps = 0.37$, we gather $10K$ gradients samples using kernel-$\epsvec$ and SL likelihoods.  These gradient estimate densities are shown in Figure~\ref{fig:exp-varg}.  An unbiased estimate of the gradient should be centered at $0$.  There are two important results.  First, the SL estimates have a small bias, even at $S=50$.  This is because it is estimating the true Gamma distribution of $\pi(\x|\thetav)$ with a Gaussian.  We can analytically estimate this bias as $S \rightarrow \infty$; for this example it is $-7.8$ which is what SL estimates are centered around ($-9.3$ for $S=5$ and $7.3$ for $S=50$).  The kernel-$\epsvec$ likelihood, on the other hand, exhibits low bias at $S=50$.  However, the second important result is the variances.  SL variances decrease quickly with $S$: $\sigma^2 = 43^2 \rightarrow 4.9^2$, whereas kernel-$\epsvec$ starts very high and remains high: $\sigma^2 = 147^2 \rightarrow 19^2$.  It is for this reason that we have chosen to use SL likelihoods for our gradient estimates, despite their small bias. As mentioned in Section~\ref{sec:habc-grads} it is possible that other likelihood models, such as KDE, might provide low bias and low variance gradient estimates.  We leave this for future work.
%
%

\begin{figure}[t]
\setlength{\linewidth}{\textwidth}
\setlength{\hsize}{\textwidth}
\begin{center}  
\includegraphics[width=0.45\columnwidth]{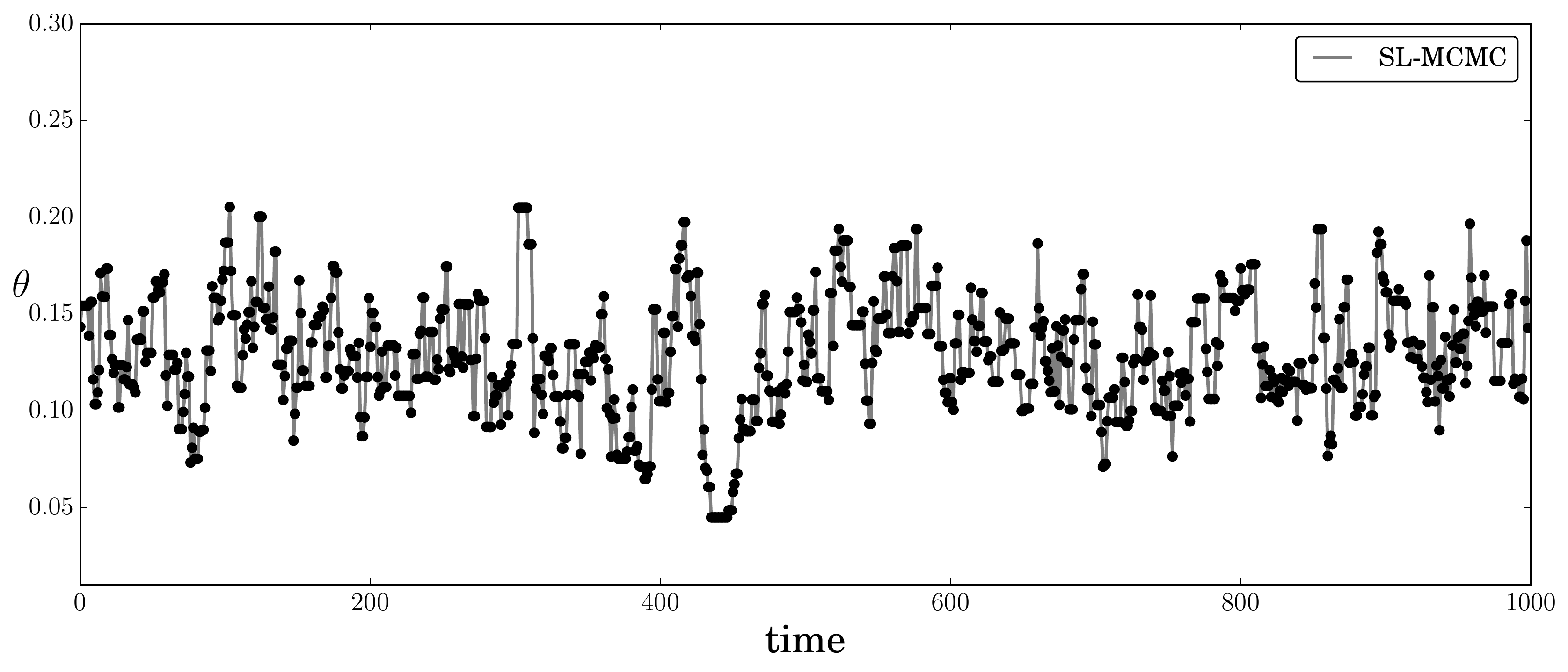}
\includegraphics[width=0.45\columnwidth]{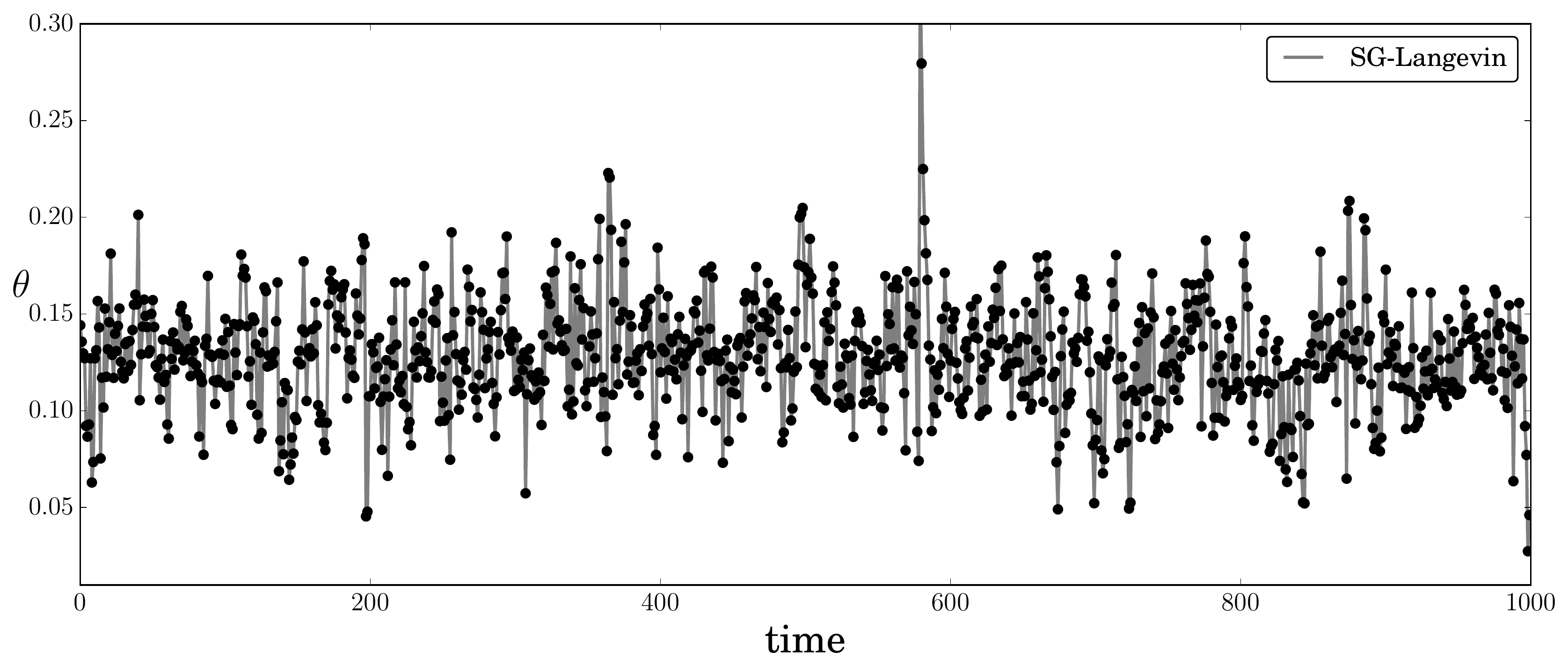}
\includegraphics[width=0.45\columnwidth]{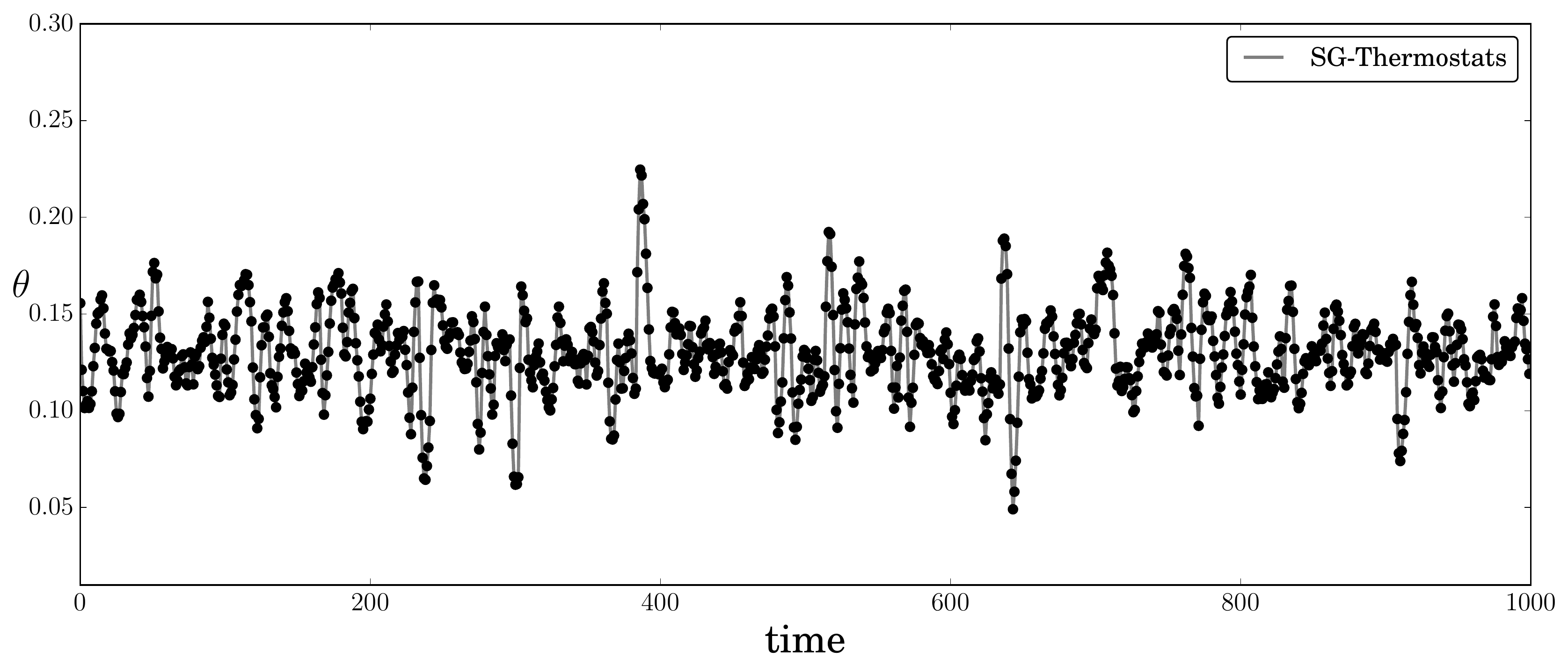}
\includegraphics[width=0.45\columnwidth]{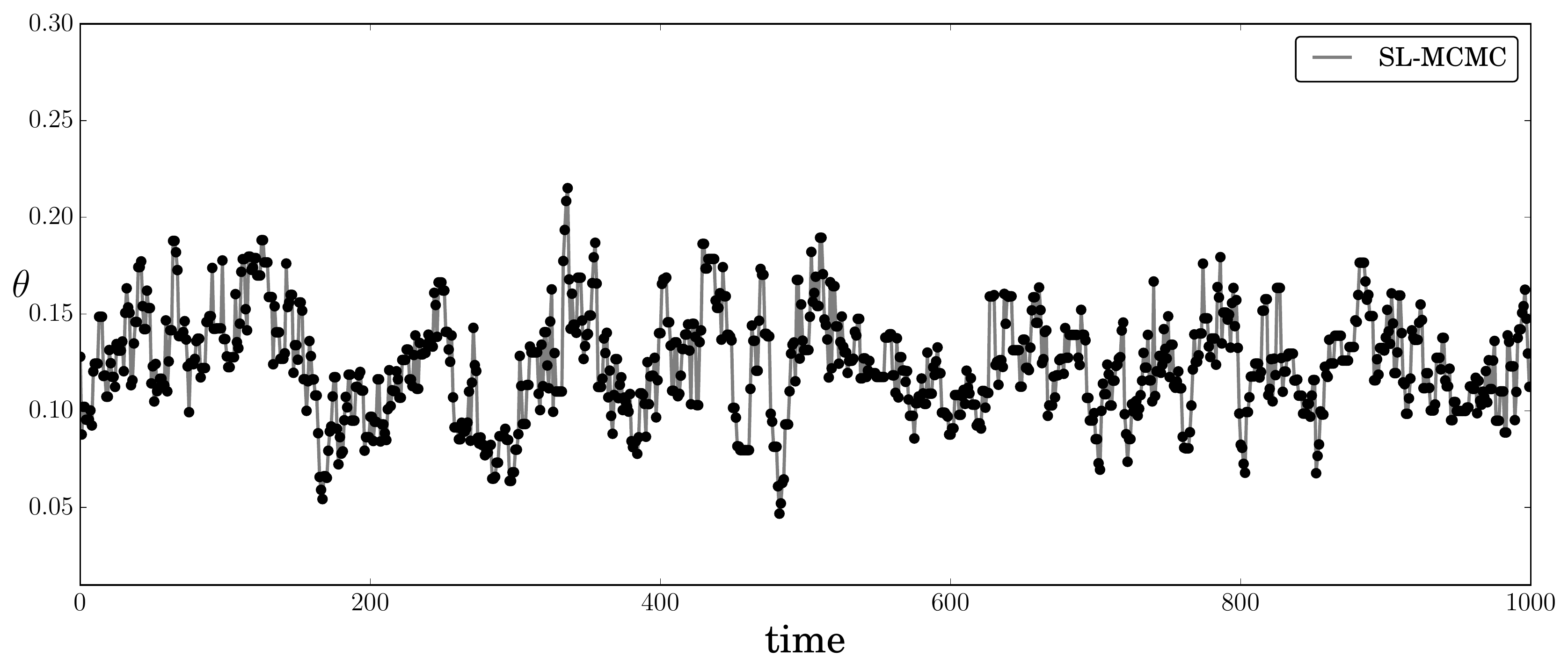}
\includegraphics[width=0.45\columnwidth]{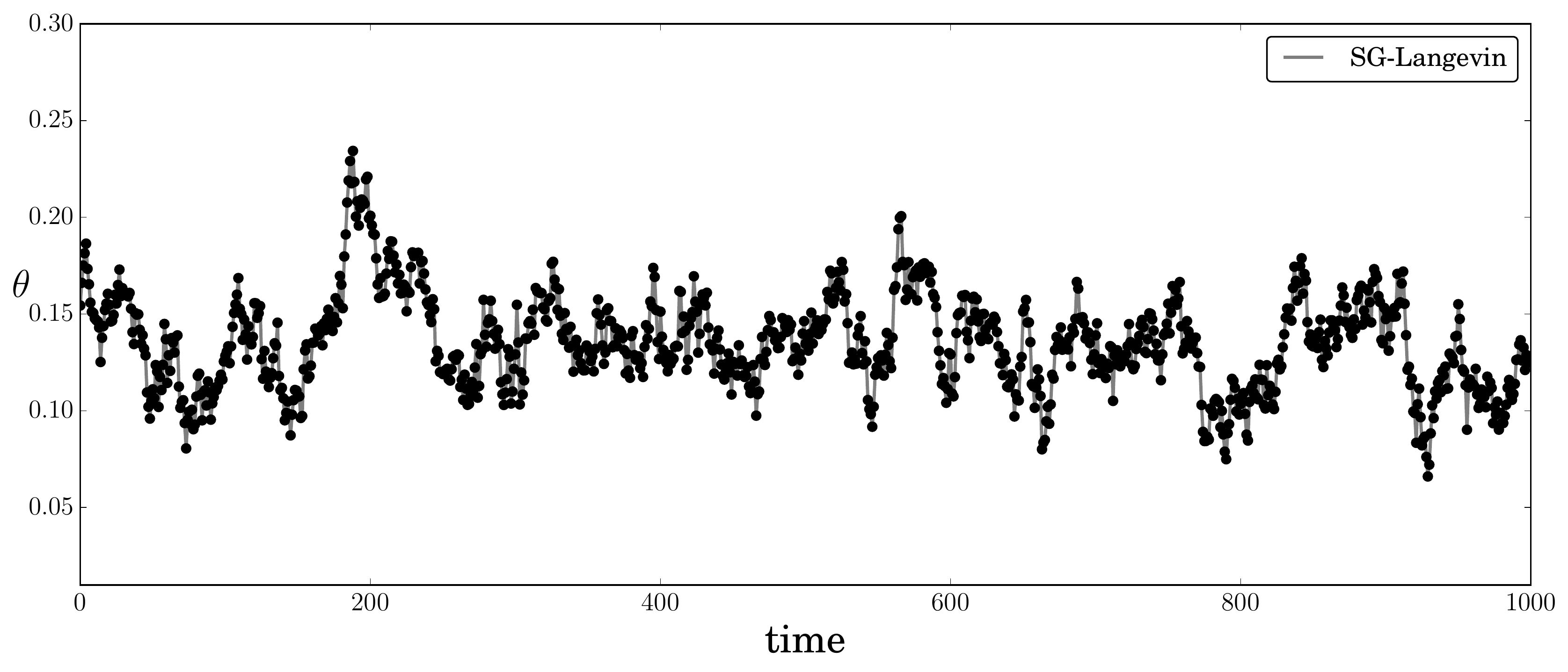}
\includegraphics[width=0.45\columnwidth]{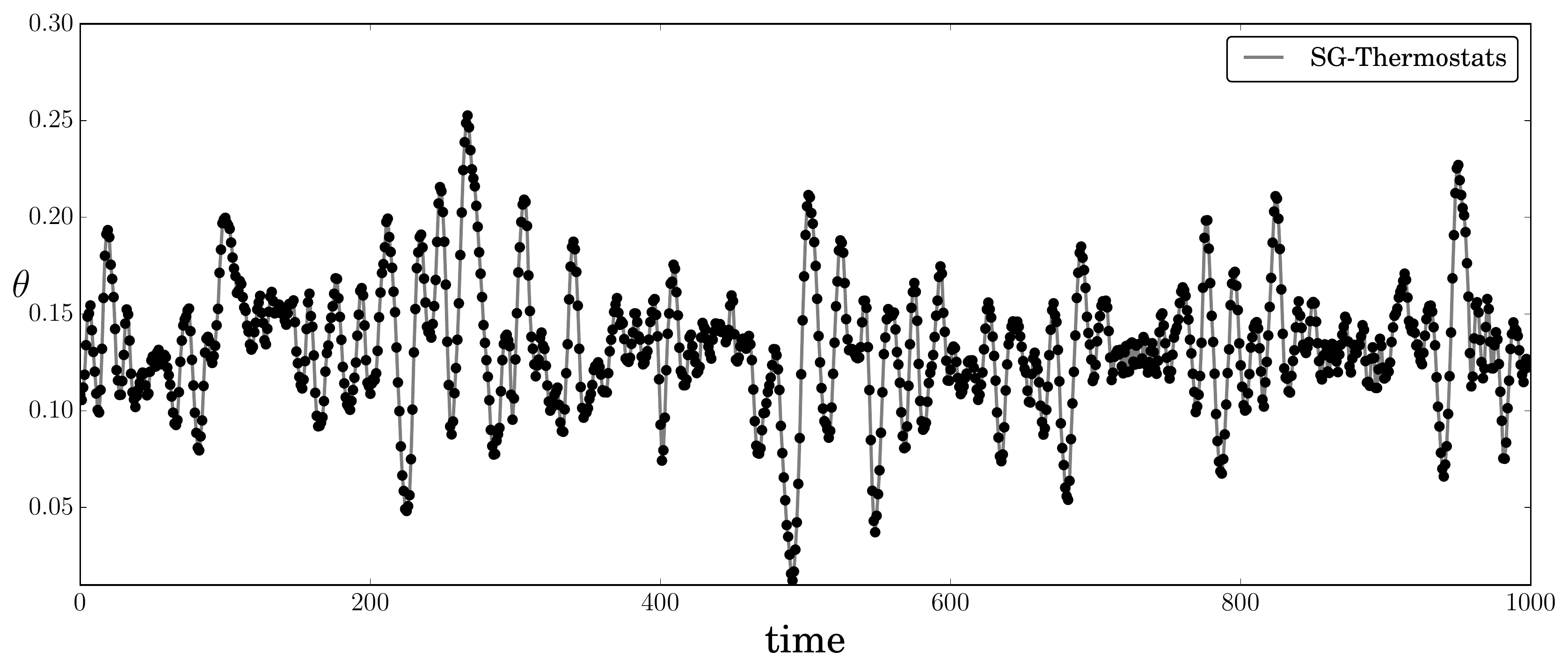}
\caption{\small{Trajectories of the last $1000$ $\thetav$ samples for the demonstration problem.  {\bf Left column:} Non-persistent random seeds.  {\bf Right column:} Persistent random seeds with $\gamma = 0.1$.  Each algorithm's parameters were optimized to minimize the total variational distance.  With persistent seeds, each algorithm's random walk behavior is suppressed.  Without persistent seeds, the optimal step-size $\eta$ for SG-Thermostats is small, resulting in an under-dispersed estimate of the posterior; when the seeds are persistent, the gradients are more consistent, and the optimal step-size is larger and therefore there is larger injected noise.  The resulting posteriors are shown in Figure~\ref{fig:exp-posteriors}.
}}
\label{fig:exp-theta-traces}
\end{center}
\end{figure}

\subsection{Posterior Inference using HABC}
We ran chains of length $50K$ for SL-MCMC, SGLD, SGHMC, and SGNHT versions of HABC using SL gradient estimates ($S=5$).  A pseudo-marginal version of SL-MCMC was used.  We note that SGHMC gave results nearly identical to SGNHT, so are not shown do to space limitations.   In one set of experiments, common random seeds were used for gradient computations only, and did not persist over time steps; these experiments are called {\em non-persistent}.  In another set of runs, we resampled $\omega_s$ at each time step with probability $\gamma = 0.1$; these experiments are {\em persistent}.  In Figure~\ref{fig:exp-posteriors} we show the posterior distributions for these experiments; in Table~\ref{tab:exp-posterior} we report the {\em total variational distance} between the true posterior and the ABC posteriors using the first $10K$ samples and after $50K$ samples (averaged over $5$ chains).  Of note is the poor approximation of SG-Thermostats when the seeds are not persistent.  By adding persistent seeds, SG-Thermostats gives similar posteriors to the other methods.

In Figure~\ref{fig:exp-theta-traces} we show the trace plots of the last $1000$ samples from a single chain for each algorithm.  In the left column, traces for non-persistent random seeds are shown, and on the right, traces for persistent seeds.  We can observe that persistent random seeds further reduces the random walk behavior of all three methods.  We also observe small improvements in total variational distance for SL-MCMC and SGLD, while SGNHT improves significantly. We find this a compelling mystery.  Is it because of the interaction between hyperparameters and stochastic gradients, or is this an artifact of this simple model?


\begin{table}[h]
\caption{Average total variational distance (tvd) for the demonstration problem.  {\em Non-persistent} used no persistent random seeds, whereas {\em Persistent} randomly proposes a new $\omega_s$ with $\gamma=0.1$. Each algorithms' parameters were optimized for minimal tvd after $10K$ samples.  The results for SGHMC (not shown) and SGNHT are nearly identical.  
}
\label{tab:exp-posterior}
\begin{center}
\begin{tabular}{l|c|c||c|c|}
  \cline{2-5}
 & \multicolumn{2}{|c||}{Non-persistent} & \multicolumn{2}{c|}{Persistent} \\
 \cline{1-5} 
\multicolumn{1}{|l|}{Algo} & $10K$ & $50K$ & $10K$ & $50K$ \\ \hline \hline 
\multicolumn{1}{|l|}{SL-ABC} & $0.047$ & $0.045$ & $0.045$ & $0.045$ \\
\multicolumn{1}{|l|}{SGLD} & $0.049$ & $0.048$ & $0.048$ & $0.043$ \\
\multicolumn{1}{|l|}{SGNHT} & $0.232$ & $0.239$ & $0.055$ & $0.051$ \\\hline 
\end{tabular}
\end{center}
\end{table}
\section{Experiments}\label{sec:experiments}
We present experimental results comparing HABC with standard ABC-MCMC for two challenging simulators.   The first is the {\em blowfly} model which uses stochastic differential equations to model possibly chaotic population dynamics \citep{wood2010statistical}.  Although it is a low-dimensional problem, the noise and chaotic behavior of the model make it challenging for gradient-based sampling.  Our second experiment applies HABC to a Bayesian logistic regression model.  Although we only use $2$ classes ($0$'s versus $1$'s), the dimensionality is very high ($D=1568$).  We show that HABC can work well despite using SPSA  gradients.
 %

\subsection{Blowfly}\label{sec:bf}
For these experiments, a simulator of adult sheep blowfly populations \citep{wood2010statistical} is used with statistics set to those from \citep{Meeds2014GpsUai}.  The observational vector $\y$ is a time-series of a fly population counted daily. The population dynamics are modeled using a stochastic differential equation\footnote{Equation~1 in Section 1.2.3 of the supplementary information in \citep{wood2010statistical}.}
\begin{equation}
N_{t+1} = P N_{t-\tau} \exp(-N_{t-\tau}/N_0) e_t + N_t \exp(-\delta \epsilon_t) \nonumber
\end{equation}
where $e_t \sim  \mathcal{G}( 1/{\sigma_p^2},1/{\sigma_p^2})$ and $\epsilon_t 
 \sim  \mathcal{G}( 1/{\sigma_d^2},1/{\sigma_d^2})$  
are sources of noise, and $\tau$ is an integer.  In total, there are $D=6$ parameters $\theta = \{ \log P, \log \delta, \log N_0, \log \sigma_d, \log \sigma_p, \tau\}$.  As \citep{Meeds2014GpsUai} we place broad log-normal priors over $\theta_{1\ldots 5}$ and a Poisson prior over $\tau$.  This is considered a challenging problem because slight changes to some parameter settings can produce degenerate $\x$, while others settings can be very noisy due to the chaotic nature of the equations.  The statistics from \citep{Meeds2014GpsUai} are used ($J=10$): the log average of $4$ quantiles of $N/1000$, the average of $4$ quantiles of the first-order differences in $N/1000$, and the number of maximal population peaks under two different thresholds. 

We compare difference HABC algorithms with ABC-MCMC for the blowfly population problem.  We use $\epsvec = \{ 1/2,1/2,1/2,1/2, 1/4,1/4,1/4,1/4,3/4,3/4 \}$ (slightly different $\epsvec$ from \citep{Meeds2014GpsUai}) and $S=10$ for all experiments.  We use SPSA with $R=2$ using SL log-likelihoods for all HABC gradient estimates.  Without persistent seeds, the number of simulations per time-step is $2 S R$ (about double marginal ABC-MCMC) and with it is $2SR+2S\gamma$.   

\begin{figure}[ht!]
  \setlength{\linewidth}{\textwidth}
  \setlength{\hsize}{\textwidth}
  \begin{center}
    \includegraphics[width=0.45\columnwidth]{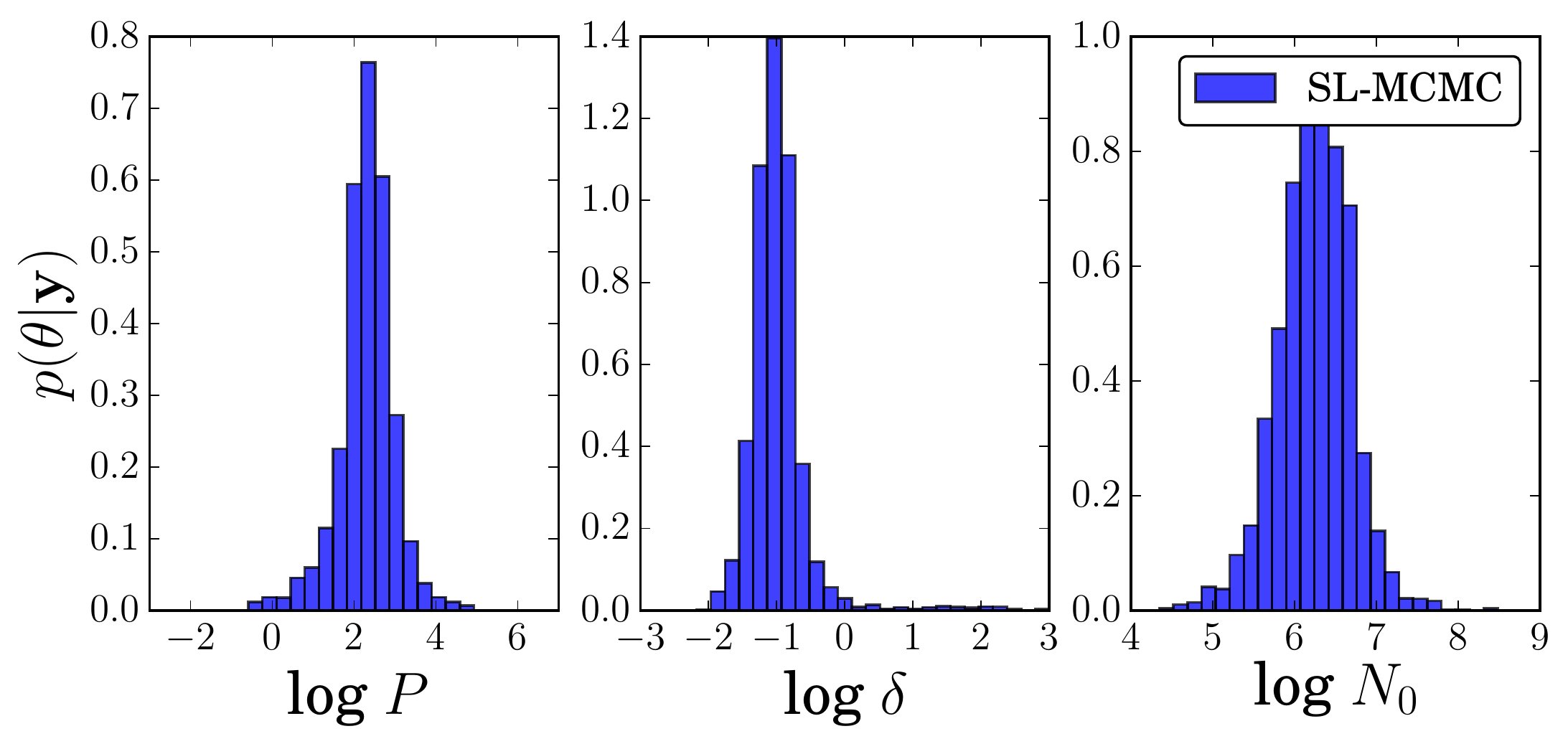}
    \includegraphics[width=0.45\columnwidth]{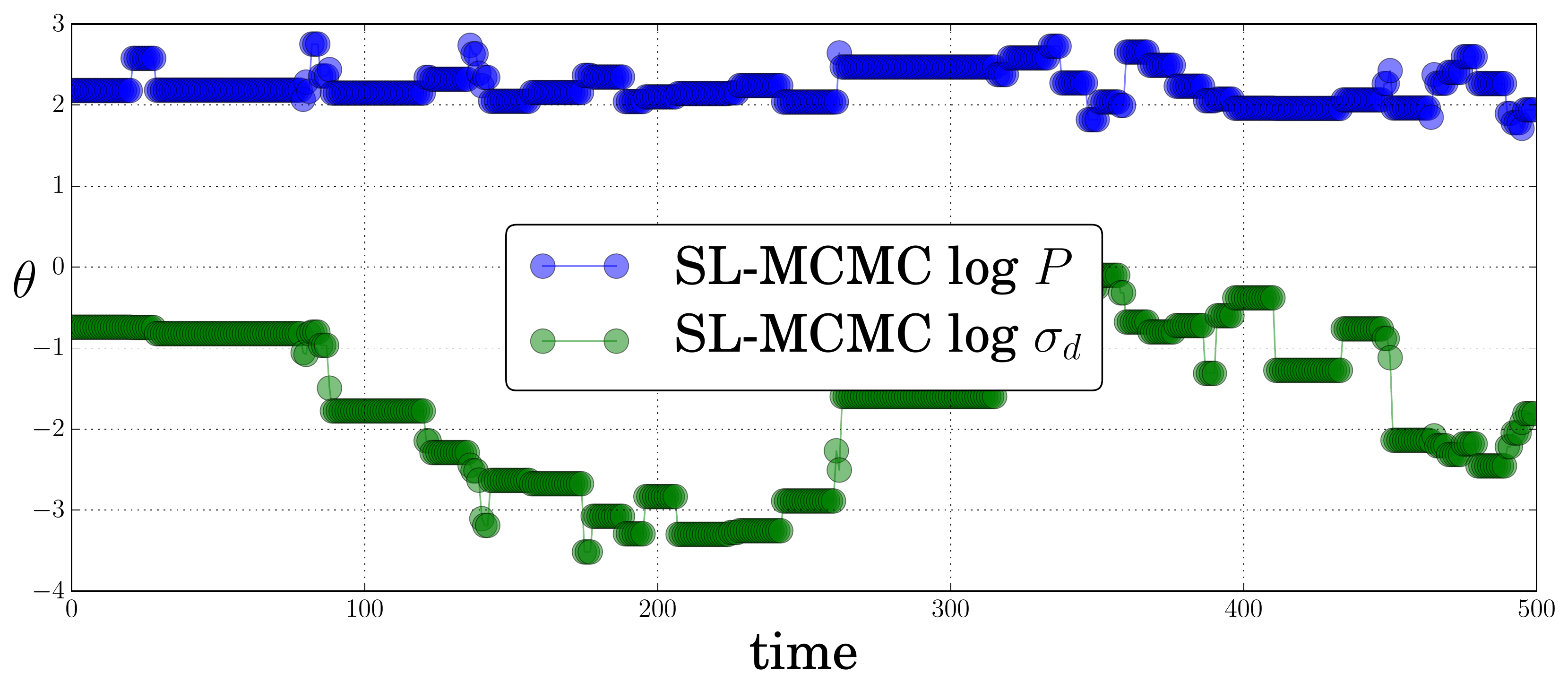}
    \includegraphics[width=0.45\columnwidth]{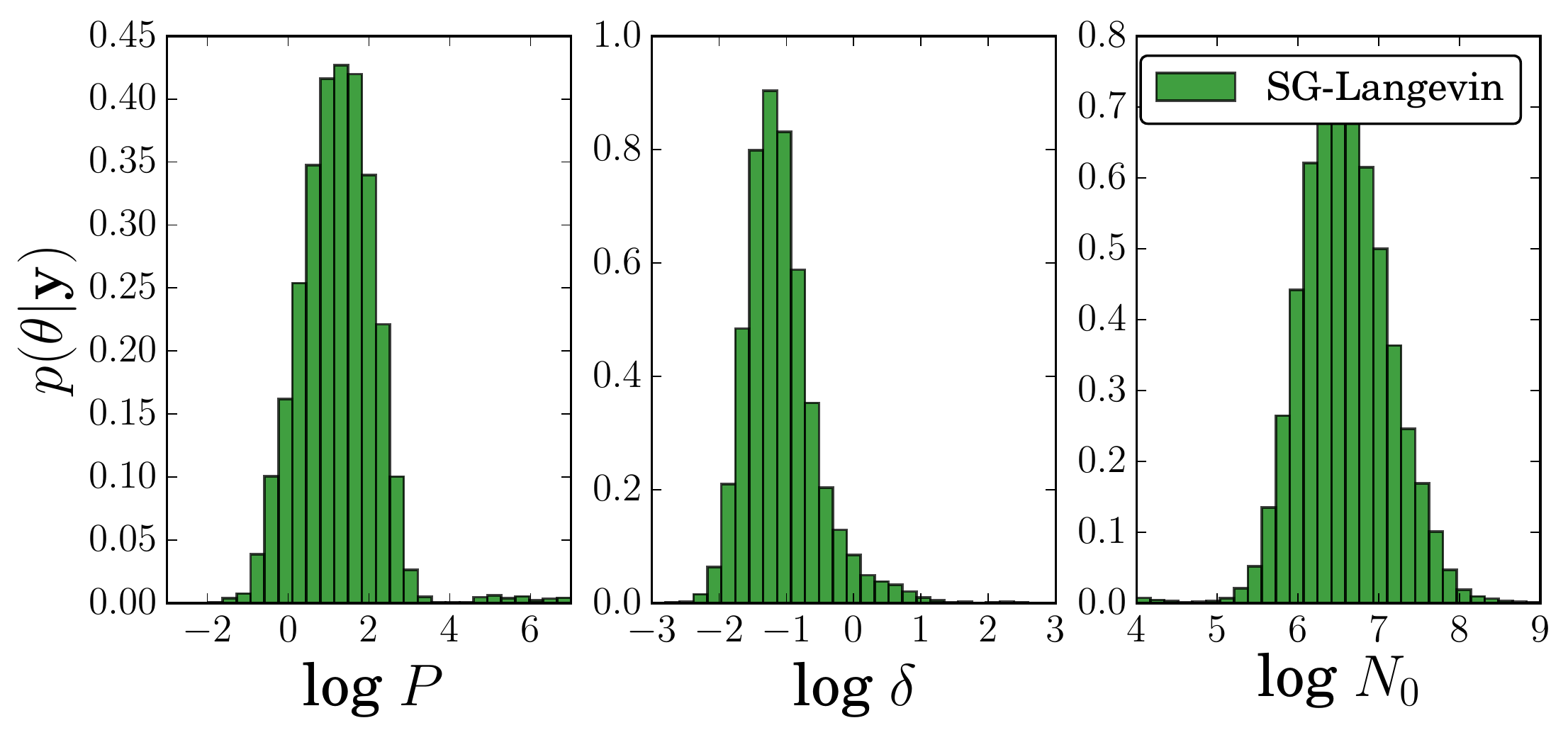}
    \includegraphics[width=0.45\columnwidth]{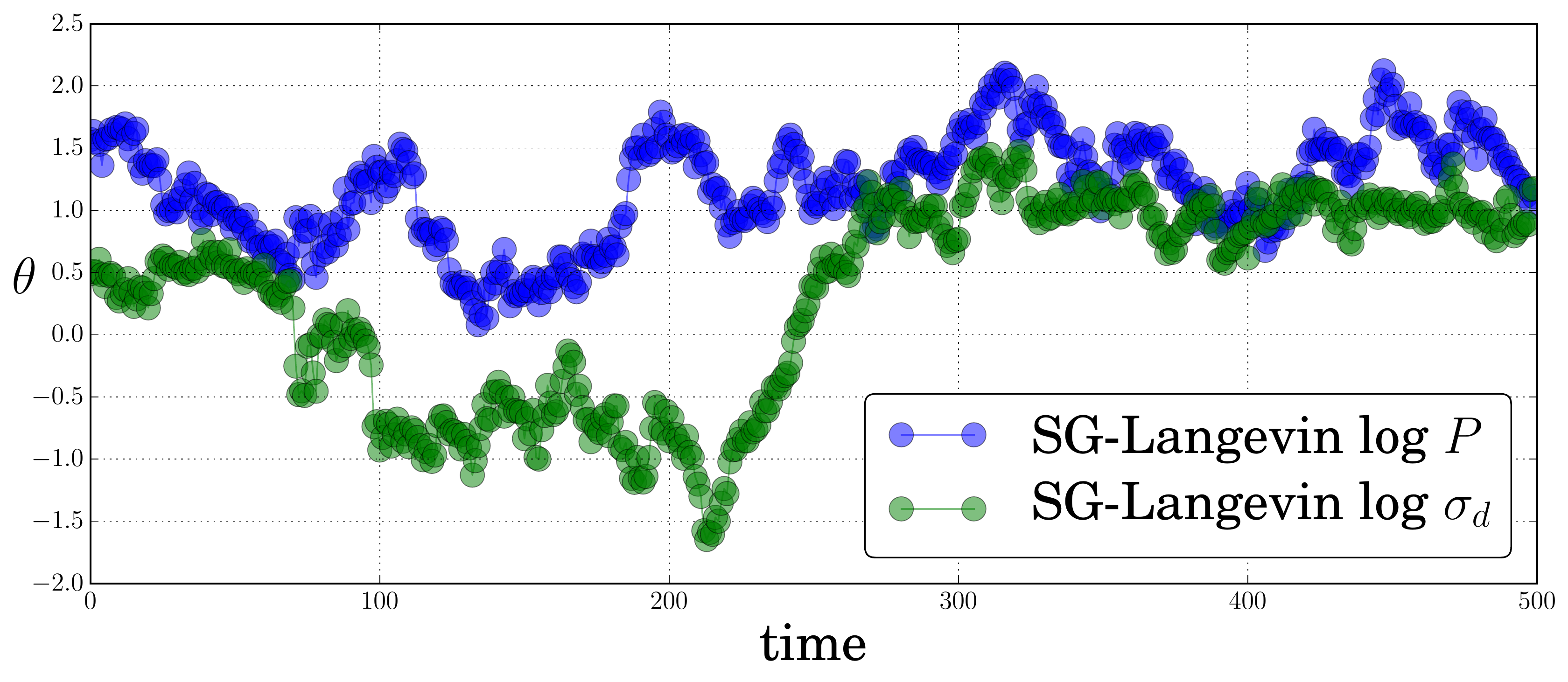}
    \includegraphics[width=0.45\columnwidth]{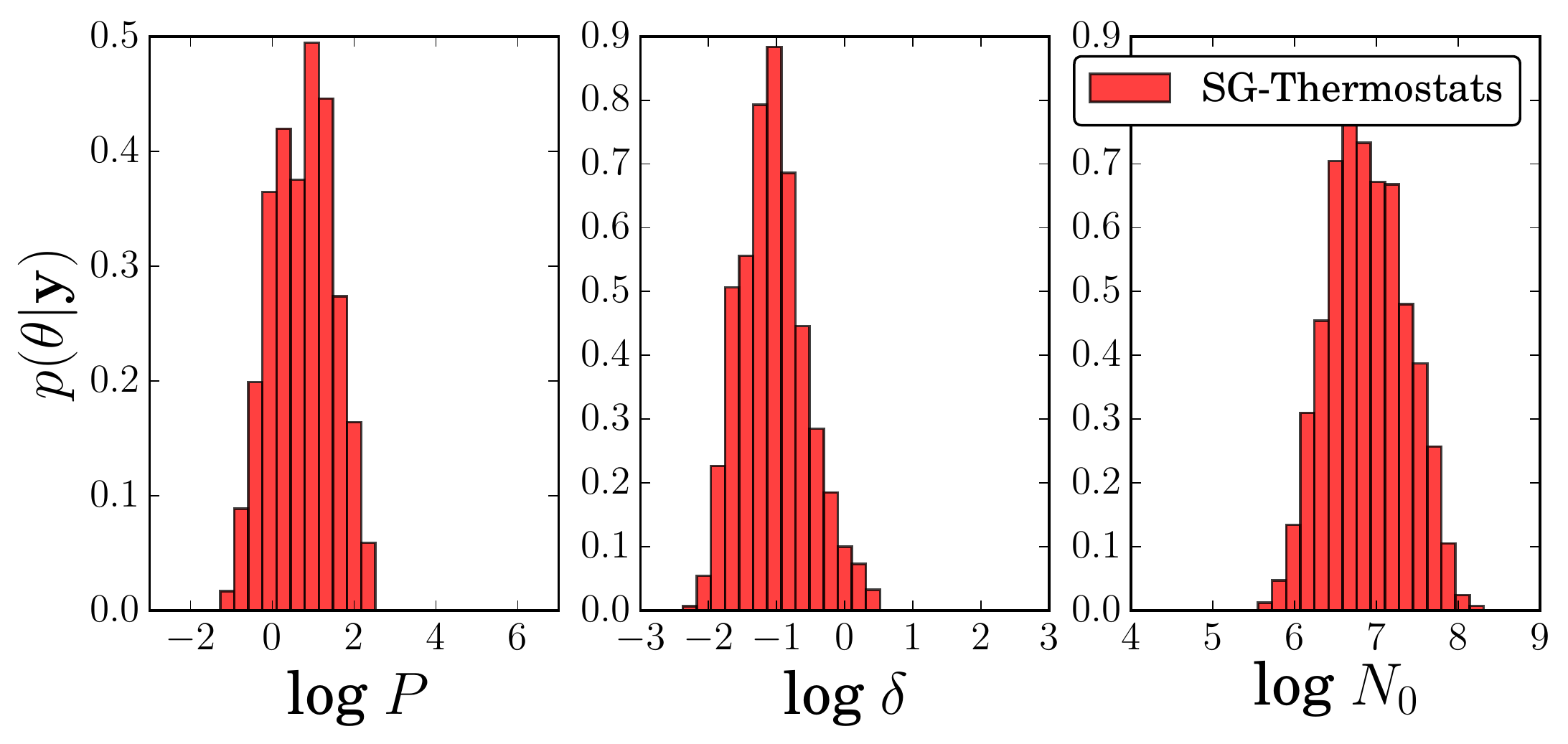}
    \includegraphics[width=0.45\columnwidth]{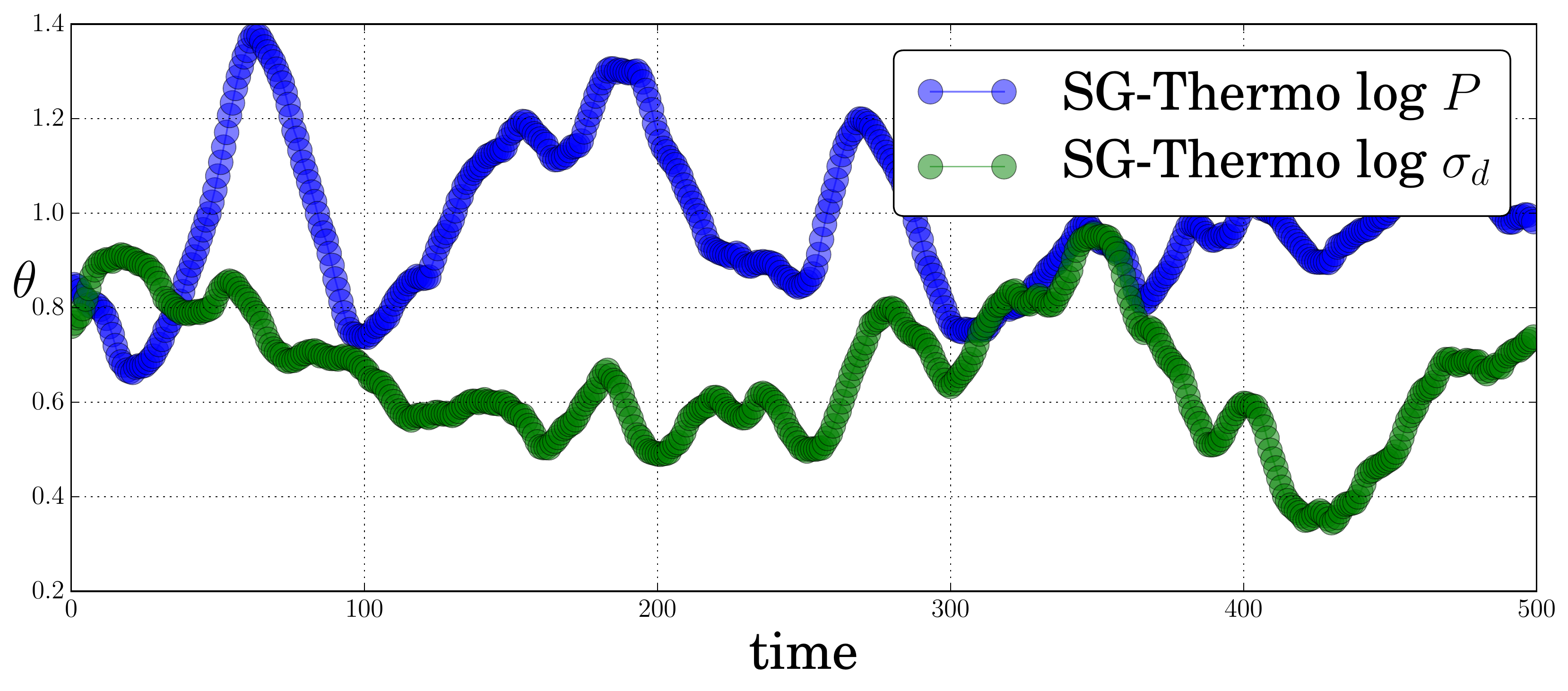}
   \caption{\small{Blowfly posterior distributions (non-persistent seeds).  {\bf Left column}:  Posteriors for three parameters for SL-MCMC (top row), SGLD (middle), and SG-Thermostats (bottom).  {\bf Right column:} Trajectories of the last 1000 samples for two parameters.
  }}
  \label{fig:bf-results}
  \end{center}
  \end{figure}

\begin{figure}[t]
\begin{center}
\includegraphics[width=0.31\columnwidth]{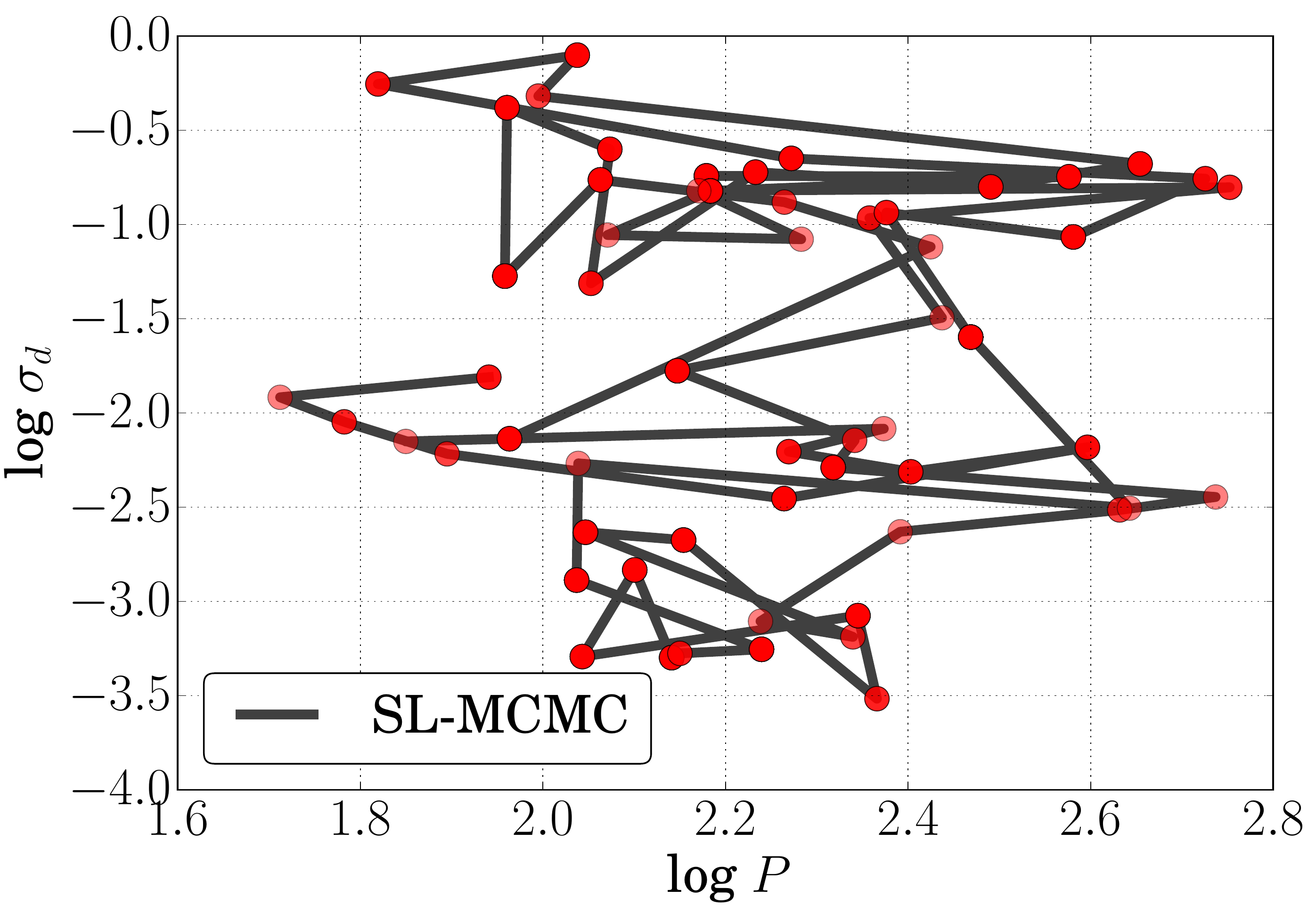}
\includegraphics[width=0.31\columnwidth]{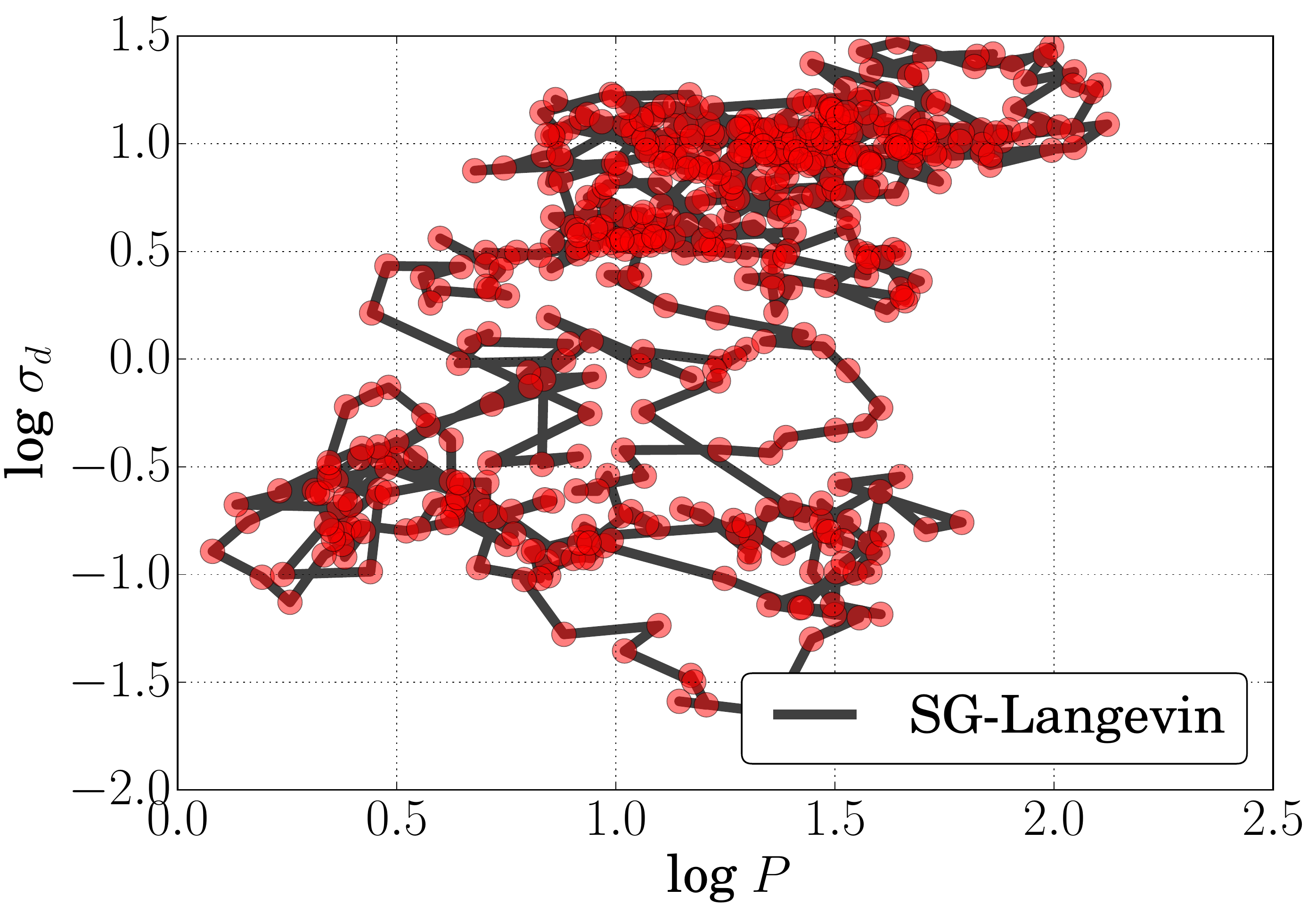}
\includegraphics[width=0.31\columnwidth]{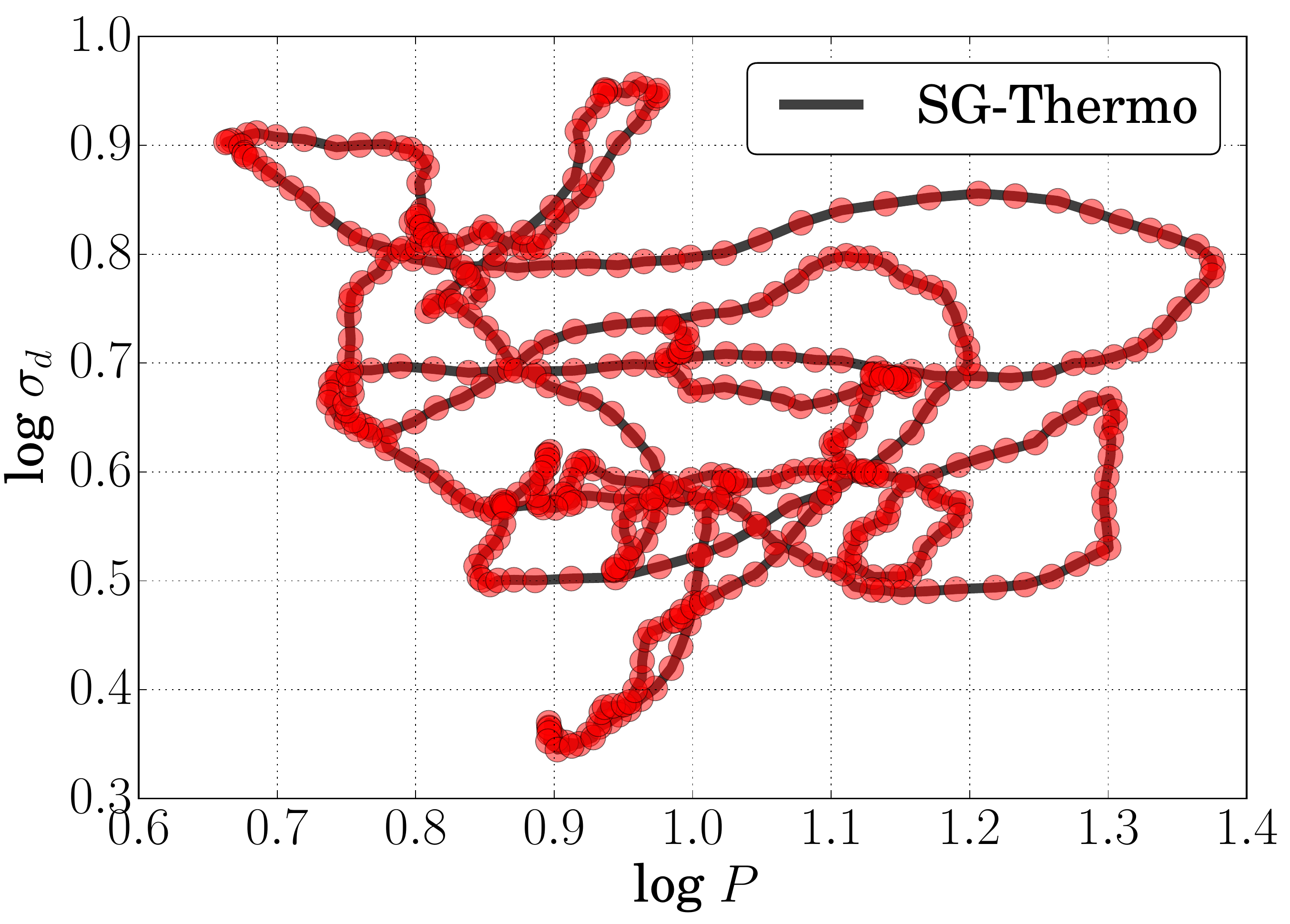}
\caption{\small{Blowfly trajectories of two parameters over the last 1000 time-steps.  {\bf Left}: SL-MCMC,  {\bf Middle:} SGLD and {\bf Right}: SG-Thermostats.  Relative to SL-MCMC, the Hamiltonian dynamics clearly show persistent $\thetav$ trajectories.}}
\label{fig:bf-two-d-theta}
\end{center}
\end{figure}

Figure~\ref{fig:bf-results} show the posterior distributions for three parameters for SL-MCMC, SGLD, and SG-Thermostats using non-persistent seeds (persistent seeds, not shown, produced very similar posteriors).  In the second row we show the trajectories of two parameters, clearly showing the suppressed random walk behavior of SGLD and SG-Thermostats relative to ABC-MCMC.  In Figure~\ref{fig:bf-two-d-theta} the scatter plots of trajectories are shown for two parameters.  
Though not shown due to space limitations, we have found that persistent seeds can improve convergence of the posterior predictive distribution.  Further experiments with persistent seeds needs to be carried out to understand the extent to which the help and how to determine when they are necessary, if at all.

  
%

\subsection{Bayesian Logistic Regression}\label{sec:auto}
\begin{figure}[t]
\begin{center}
\includegraphics[width=0.45\columnwidth]{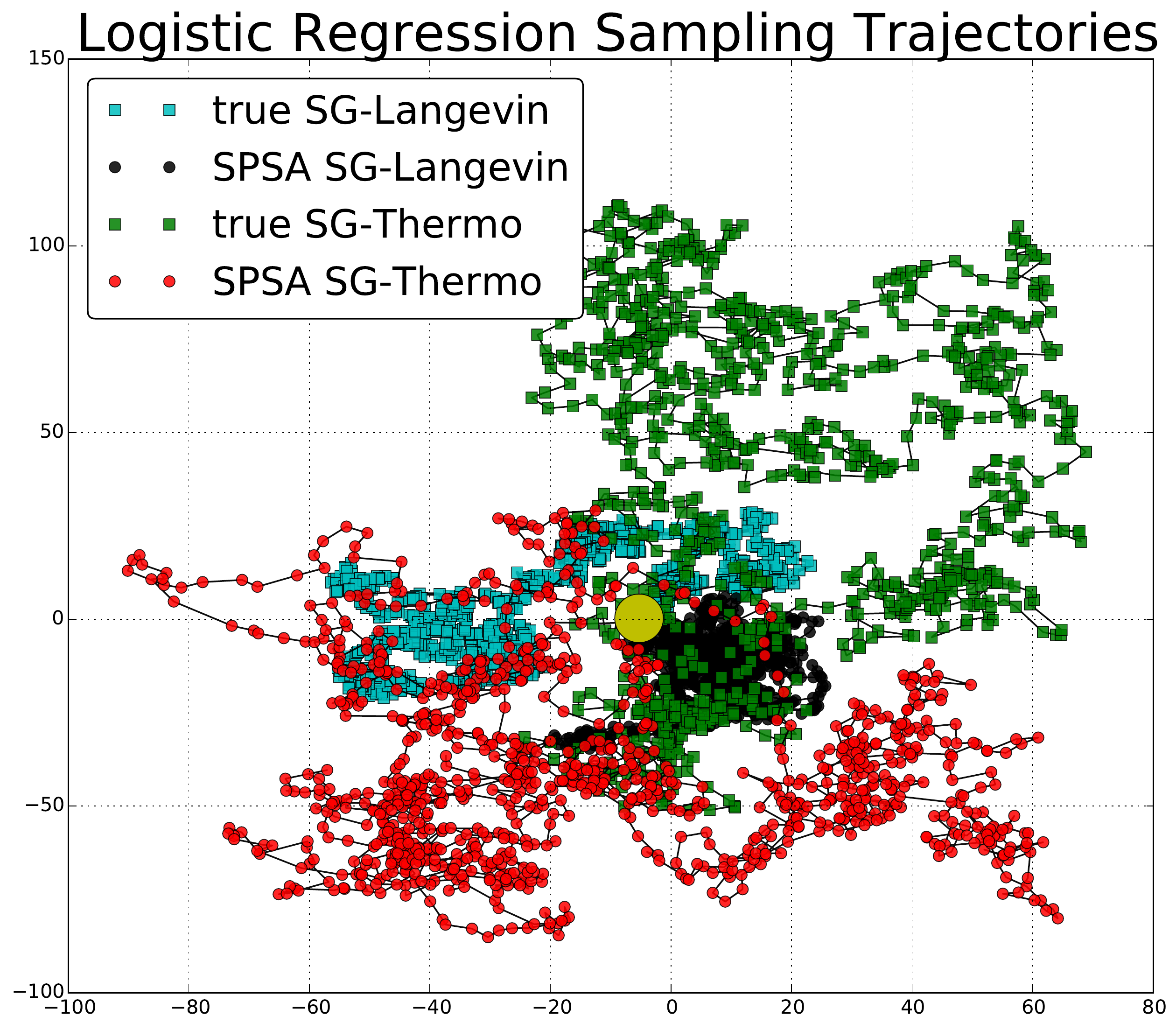}
\caption{\small{Bayesian logistic regression sampling trajectories randomly projected.  The yellow circle is the projected $\thetav_{\text{MAP}}$.}  
}
\label{fig:blr}
\end{center}
\end{figure}

We perform Bayesian inference on Bayesian inference on a logistic regression model using the digits $0$ and $1$ from MNIST.  Despite its simplicity, the model still represents a high-dimensional problem for HABC ($D=1568$).  We first ran stochastic gradient descent to determine $\thetav_{\text{MAP}}$ using the true gradient.  We then run HABC SGLD and SG-Thermostats starting  $\thetav_{\text{MAP}}$ to discover how well the algorithms explore the posterior.  We compare with SGLD and SG-Thermostats using the {\em true} gradients.   
We use $n=100$ size mini-batches and $R=10$ number of perturbations for SPSA.  Figure~\ref{fig:blr} shows samples randomly projected onto 2 dimensions ($1000$ evenly sub-sampled from $10K$).  We can see that the trajectories using SPSA exhibit similar behavior to Bayesian learning with the true gradients.  This is a positive result that indicates HABC can successfully exploit the noisy and less informative gradients of SPSA.

%



\section{DISCUSSION AND CONCLUSION} \label{sec:conclusion}

Hamiltonian ABC proposes a new set of algorithms for Bayesian inference of likelihood-free models.  HABC builds  upon the connections between Hamiltonian Monte Carlo with stochastic gradients and well-established gradient approximations based on a minimal number of forward simulations, even in high-dimensions.  We have performed some preliminary experiments showing the feasibility of running ABC on both small and large problems, and we hope that the door has been opened for exploration of larger simulation-based models using HABC. 

Another innovation we introduce is the use of persistent random seeds to suppress the simulator noise and therefore smooth the simulation landscape over a local region of parameter space. For some algorithms run on certain models, improved performance has been observed.  This is most likely to be the case for simulators with large additive noise and algorithms that benefit from long Hamiltonian trajectories (i.e. SGHMC and SG-Thermostats).  We feel that new classes of ABC algorithms could develop from using persistent random seeds, not just gradient-based samplers but traditional ABC-MCMC.
 
There are several unresolved and open questions regarding the application of stochastic gradients to ABC.  The first issue is the importance of the bias-variance relationship for different ABC likelihood models.    We found that using gradients based on the synthetic-likelihood greatly reduced their variance, but introduced a small bias, because of its Gaussian assumption.  The second issue is setting algorithm parameters, in particular the step-sizes $\eta$, the injected noise $C$ (for SGHMC/SGNHT), and the number of SPSA repetitions $R$.  All of these parameters are highly interactive.  Can we use statistical tests during the MCMC run to determine $R$?  Should $\eta$ and $C$ be set differently in the ABC setting?  One final issue is monitoring or determining whether the correct amount of noise is being injected to ensure proper sampling.  In SGLD \citep{welling2011bayesian}, for example, we can always turn down $\eta$ so that the injected noise term dominates, but when our goal is efficient exploration of the posterior, this is  not a very satisfying solution.


Expensive simulators are an important class of models that we do not address in this work.  However, previous work in Bayesian inference has shown the usefulness of HMC-based proposals based on Gaussian process of log-likelihood surfaces \citep{rasmussen:2003}.   We could similarly use HABC with  ABC surrogate models \citep{Meeds2014GpsUai,wilkinson:2014} to minimize simulation calls, yet still  benefit from Hamiltonian dynamics.  

\clearpage
\bibliographystyle{icml2014}
\bibliography{abcsgld}

\begin{thebibliography}{27}
\providecommand{\natexlab}[1]{#1}
\providecommand{\url}[1]{\texttt{#1}}
\expandafter\ifx\csname urlstyle\endcsname\relax
  \providecommand{\doi}[1]{doi: #1}\else
  \providecommand{\doi}{doi: \begingroup \urlstyle{rm}\Url}\fi

\bibitem[Andrieu \& Roberts(2009)Andrieu and Roberts]{andrieu2009pseudo}
Andrieu, C. and Roberts, G.
\newblock The pseudo-marginal approach for efficient monte carlo computations.
\newblock \emph{The Annals of Statistics}, 37\penalty0 (2):\penalty0 697--725,
  2009.

\bibitem[Beaumont et~al.(2002)Beaumont, Zhang, and
  Balding]{beaumont2002approximate}
Beaumont, Mark~A, Zhang, Wenyang, and Balding, David~J.
\newblock Approximate bayesian computation in population genetics.
\newblock \emph{Genetics}, 162\penalty0 (4):\penalty0 2025--2035, 2002.

\bibitem[Chen et~al.(2014)Chen, Fox, and Guestrin]{chen2014stochastic}
Chen, Tianqi, Fox, Emily~B, and Guestrin, Carlos.
\newblock Stochastic gradient hamiltonian monte carlo.
\newblock 2014.

\bibitem[Ding et~al.(2014)Ding, Fang, Babbush, Chen, Skeel, and
  Neven]{ding2014bayesian}
Ding, Nan, Fang, Youhan, Babbush, Ryan, Chen, Changyou, Skeel, Robert~D, and
  Neven, Hartmut.
\newblock Bayesian sampling using stochastic gradient thermostats.
\newblock In \emph{Advances in Neural Information Processing Systems}, pp.\
  3203--3211, 2014.

\bibitem[Duane et~al.(1987)Duane, Kennedy, Pendleton, and
  Roweth]{duane1987hybrid}
Duane, Simon, Kennedy, Anthony~D, Pendleton, Brian~J, and Roweth, Duncan.
\newblock Hybrid monte carlo.
\newblock \emph{Physics letters B}, 195\penalty0 (2):\penalty0 216--222, 1987.

\bibitem[Ehrlich et~al.(2013)Ehrlich, Jasra, and Kantas]{Ehrlich2013}
Ehrlich, Elena, Jasra, Ajay, and Kantas, Nikolas.
\newblock Gradient free parameter estimation for hidden markov models with
  intractable likelihoods.
\newblock \emph{Methodology and Computing in Applied Probability}, pp.\  1--35,
  2013.

\bibitem[Fan et~al.(2013)Fan, Nott, and Sisson]{fan:2013}
Fan, Yanan, Nott, David~J, and Sisson, Scott~A.
\newblock Approximate bayesian computation via regression density estimation.
\newblock \emph{Stat}, 2013.

\bibitem[Kiefer et~al.(1952)Kiefer, Wolfowitz, et~al.]{kiefer1952stochastic}
Kiefer, Jack, Wolfowitz, Jacob, et~al.
\newblock Stochastic estimation of the maximum of a regression function.
\newblock \emph{The Annals of Mathematical Statistics}, 23\penalty0
  (3):\penalty0 462--466, 1952.

\bibitem[Kleinman et~al.(1999)Kleinman, Spall, and
  Naiman]{kleinman1999simulation}
Kleinman, Nathan~L, Spall, James~C, and Naiman, Daniel~Q.
\newblock Simulation-based optimization with stochastic approximation using
  common random numbers.
\newblock \emph{Management Science}, 45\penalty0 (11):\penalty0 1570--1578,
  1999.

\bibitem[Leimkuhler \& Reich(2009)Leimkuhler and
  Reich]{leimkuhler2009metropolis}
Leimkuhler, Benedict and Reich, Sebastian.
\newblock A metropolis adjusted nos{\'e}-hoover thermostat.
\newblock \emph{ESAIM: Mathematical Modelling and Numerical Analysis},
  43\penalty0 (04):\penalty0 743--755, 2009.

\bibitem[Marin et~al.(2012)Marin, Pudlo, Robert, and Ryder]{marin:2012}
Marin, J.-M., Pudlo, P., Robert, C.P., and Ryder, R.J.
\newblock Approximate bayesian computational methods.
\newblock \emph{Statistics and Computing}, 22:\penalty0 1167--1180, 2012.

\bibitem[Marjoram et~al.(2003)Marjoram, Molitor, Plagnol, and
  Tavar{\'e}]{marjoram2003markov}
Marjoram, Paul, Molitor, John, Plagnol, Vincent, and Tavar{\'e}, Simon.
\newblock Markov chain monte carlo without likelihoods.
\newblock \emph{Proceedings of the National Academy of Sciences}, 100\penalty0
  (26):\penalty0 15324--15328, 2003.

\bibitem[Meeds \& Welling(2014)Meeds and Welling]{Meeds2014GpsUai}
Meeds, Edward and Welling, Max.
\newblock {GPS-ABC}: Gaussian process surrogate approximate bayesian
  computation.
\newblock \emph{Uncertainty in AI}, 2014.

\bibitem[Murray \& Elliott(2012)Murray and Elliott]{Murray2012}
Murray, Iain and Elliott, Lloyd~T.
\newblock Driving markov chain monte carlo with a dependent random stream.
\newblock \emph{arXiv:1204.3187}, 2012.

\bibitem[Neal(2011)]{neal2011mcmc}
Neal, Radford~M.
\newblock Mcmc using hamiltonian dynamics.
\newblock \emph{Handbook of Markov Chain Monte Carlo}, 2, 2011.

\bibitem[Neal(2012)]{Neal2012}
Neal, Radford~M.
\newblock How to view an mcmc simulation as a permutation, with applications to
  parallel simulation and improved importance sampling.
\newblock \emph{Technical Report No. 1201, Dept. of Statistics, University of
  Toronto}, 2012.

\bibitem[Rasmussen(2003)]{rasmussen:2003}
Rasmussen, C.E.
\newblock Gaussian processes to speed up hybrid monte carlo for expensive
  bayesian integrals.
\newblock \emph{Bayesian Statistics}, 7:\penalty0 651--659, 2003.

\bibitem[Sisson et~al.(2007)Sisson, Fan, and Tanaka]{sisson2007sequential}
Sisson, SA, Fan, Y, and Tanaka, Mark~M.
\newblock Sequential monte carlo without likelihoods.
\newblock \emph{Proceedings of the National Academy of Sciences}, 104\penalty0
  (6):\penalty0 1760, 2007.

\bibitem[Sisson \& Fan(2010)Sisson and Fan]{sisson:2010}
Sisson, Scott~A and Fan, Yanan.
\newblock Likelihood-free markov chain monte carlo.
\newblock \emph{Arxiv preprint arXiv:1001.2058}, 2010.

\bibitem[Spall(1992)]{spall1992multivariate}
Spall, James~C.
\newblock Multivariate stochastic approximation using a simultaneous
  perturbation gradient approximation.
\newblock \emph{Automatic Control, IEEE Transactions on}, 37\penalty0
  (3):\penalty0 332--341, 1992.

\bibitem[Spall(2000)]{spall2000adaptive}
Spall, James~C.
\newblock Adaptive stochastic approximation by the simultaneous perturbation
  method.
\newblock \emph{Automatic Control, IEEE Transactions on}, 45\penalty0
  (10):\penalty0 1839--1853, 2000.

\bibitem[Spall(2005)]{spall2005monte}
Spall, James~C.
\newblock Monte carlo computation of the fisher information matrix in
  nonstandard settings.
\newblock \emph{Journal of Computational and Graphical Statistics}, 14\penalty0
  (4), 2005.

\bibitem[Turner \& Sederberg(2014)Turner and Sederberg]{TurnerGenLik2014}
Turner, Brandon~M. and Sederberg, Per~B.
\newblock A generalized, likelihood-free method for posterior estimation.
\newblock \emph{Psychonomic Bulletin \& Review}, 21\penalty0 (2):\penalty0
  227--250, 2014.

\bibitem[Welling \& Teh(2011)Welling and Teh]{welling2011bayesian}
Welling, Max and Teh, Yee~W.
\newblock Bayesian learning via stochastic gradient langevin dynamics.
\newblock In \emph{Proceedings of the 28th International Conference on Machine
  Learning (ICML-11)}, pp.\  681--688, 2011.

\bibitem[Wilkinson(2013)]{Wilkinson2013}
Wilkinson, R.
\newblock Approximate bayesian computation ({ABC}) gives exact results under
  the assumption of model error.
\newblock \emph{Statistical Applications in Genetics and Molecular Biology},
  12\penalty0 (2):\penalty0 129--142, 2013.

\bibitem[Wilkinson(2014)]{wilkinson:2014}
Wilkinson, R.
\newblock Accelerating abc methods using gaussian processes.
\newblock \emph{AISTATS}, 2014.

\bibitem[Wood(2010)]{wood2010statistical}
Wood, Simon~N.
\newblock Statistical inference for noisy nonlinear ecological dynamic systems.
\newblock \emph{Nature}, 466\penalty0 (7310):\penalty0 1102--1104, 2010.

\end{thebibliography}

\end{document}